\documentclass[10pt,twocolumn]{article}
\newcommand{\R}{\mathbb{R}}
\usepackage[most]{tcolorbox}
\tcbuselibrary{listings, breakable}

\newtcolorbox{myequation}[1][]{colback=gray!10!white, colframe=black, 
  fonttitle=\bfseries, title=#1, sharp corners, boxrule=0.8pt}

\usepackage{multicol}
\usepackage{multirow}
\usepackage{array}
    \newcolumntype{M}[1]{>{\centering\arraybackslash}m{#1}}
\usepackage{url}
\usepackage{xcolor}
\usepackage{booktabs}       
\usepackage{amsfonts}       
\usepackage{nicefrac}       
\usepackage{microtype} 
\usepackage{array}
\usepackage{lipsum}		
\usepackage{graphicx}
\usepackage{doi}

\renewcommand{\S}{\mathcal{S}}
\usepackage{tikz, pgf} 
\usetikzlibrary{decorations.pathmorphing}

\usepackage{amsmath}
\usepackage{amsthm}

\theoremstyle{definition}

\usepackage[most]{tcolorbox}
\newtheorem{theorem}{Theorem}[section]
\newtheorem{proposition}[theorem]{Proposition}
\newtheorem{lemma}[theorem]{Lemma}

\theoremstyle{definition}

\theoremstyle{remark}
\newtheorem{remark}[theorem]{Remark}

\usepackage[most]{tcolorbox} 

\usepackage{amsmath,amssymb}
\usepackage{amsthm}
\usepackage[framemethod=tikz]{mdframed}

\usepackage[framemethod=tikz]{mdframed}

\definecolor{darkblue}{RGB}{0,0,139} 

\newenvironment{boxtheorem}[1][]{
  \refstepcounter{theorem}%
  \begin{mdframed}[
    roundcorner=5pt,
    linecolor=darkblue,
    linewidth=1pt,
    backgroundcolor=white 
  ]
  \noindent
  \textcolor{darkblue}{\bfseries Theorem~\thetheorem}%
  \ifx\relax#1\relax\else\textcolor{darkblue}{\bfseries\ (#1)}\fi
  \par\medskip 
}{
  \end{mdframed}
}

\usepackage{multicol}
\usepackage{multirow}
\usepackage{array}
    \newcolumntype{P}[1]{>{\centering\arraybackslash}p{#1}}
    \newcolumntype{M}[1]{>{\centering\arraybackslash}m{#1}}
\usepackage{caption}

\newcommand{\ones}{1}
\newcommand{\zeros}{0}

\newcommand{\diag}{\text{diag}}

\newcommand{\attn}{\mathrm{atten}}
\newcommand{\row}{\mathrm{row}}
\newcommand{\col}{\mathrm{col}}
\usepackage{xfrac}
\usepackage[margin=1in,columnsep=0.28in]{geometry}
\usepackage{newtxtext}
\usepackage{newtxmath}
\AtBeginDocument{}
\usepackage{microtype}             
\usepackage{graphicx}
\usepackage{subcaption}
\usepackage{booktabs}
\usepackage{algorithm}
\usepackage[noend]{algpseudocode}

\usepackage{natbib} 
\title{How do transformers align tokens? \\
Provable Optimal Transport with Transformers}
\author{Hadi Daneshmand \\
Department of Computer Science University of Virginia \\
\texttt{dhadi@virginia.edu}}
\date{} 

\begin{document}
\maketitle




\begin{abstract}
Despite their empirical success, the internal mechanism by which transformer models align tokens during language processing remains poorly understood. This paper provides a mechanistic and theoretical explanation of token alignment in LLMs. We first present empirical evidences showing that, in machine translation, attention weights progressively align translated word pairs across layers, closely approximating Optimal Transport  (OT) between word embeddings. Building on this observation, we prove that softmax self-attention layers can simulate gradient descent on the dual of the entropy-regularized OT problem, providing a theoretical foundation for the alignment. Our analysis yields a constructive convergence bound showing that transformer depth controls OT approximation accuracy. A direct implication is that standard transformers can sort lists of varying lengths without any parameter adjustment, up to an error term vanishing with transformers depth.
\end{abstract}

\section{Introduction} 

\begin{figure*}[t!]
    \centering
    \begin{tabular}{ M{4cm} M{4cm} M{4cm} |M{4cm}}
     \includegraphics[width=0.22\textwidth]{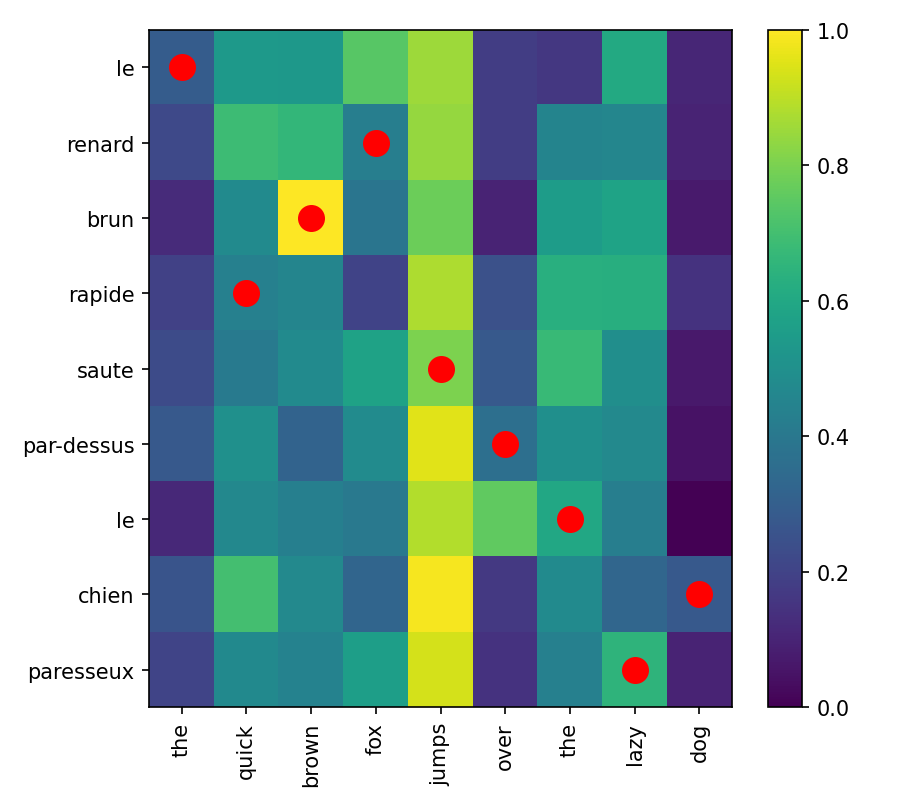}& \includegraphics[width=0.22\textwidth]{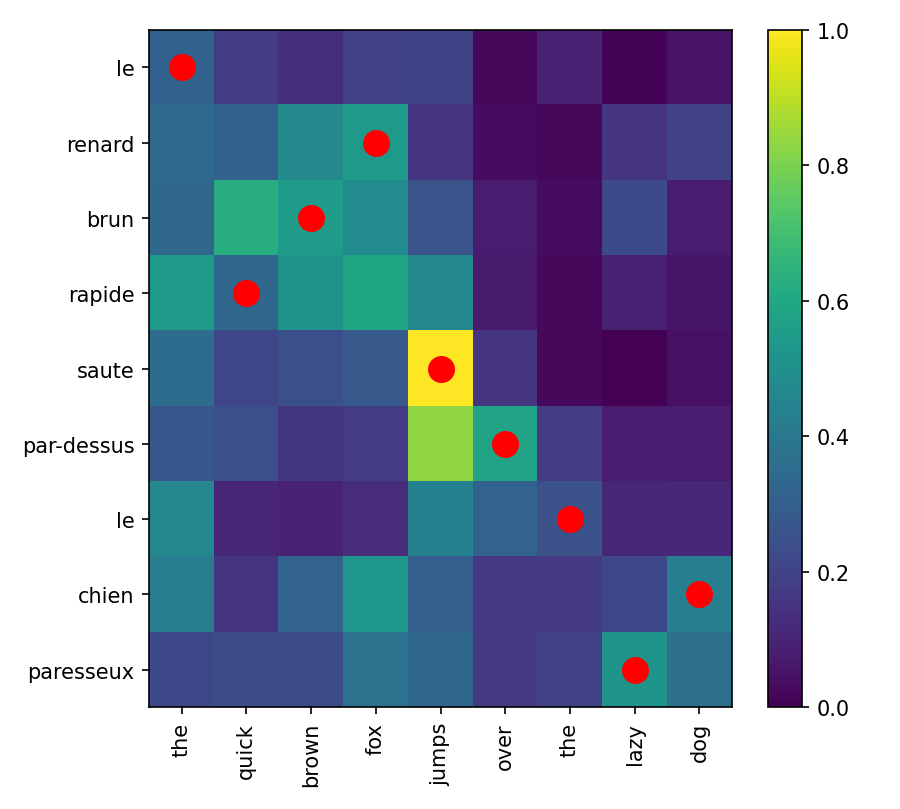} & \includegraphics[width=0.22\textwidth]{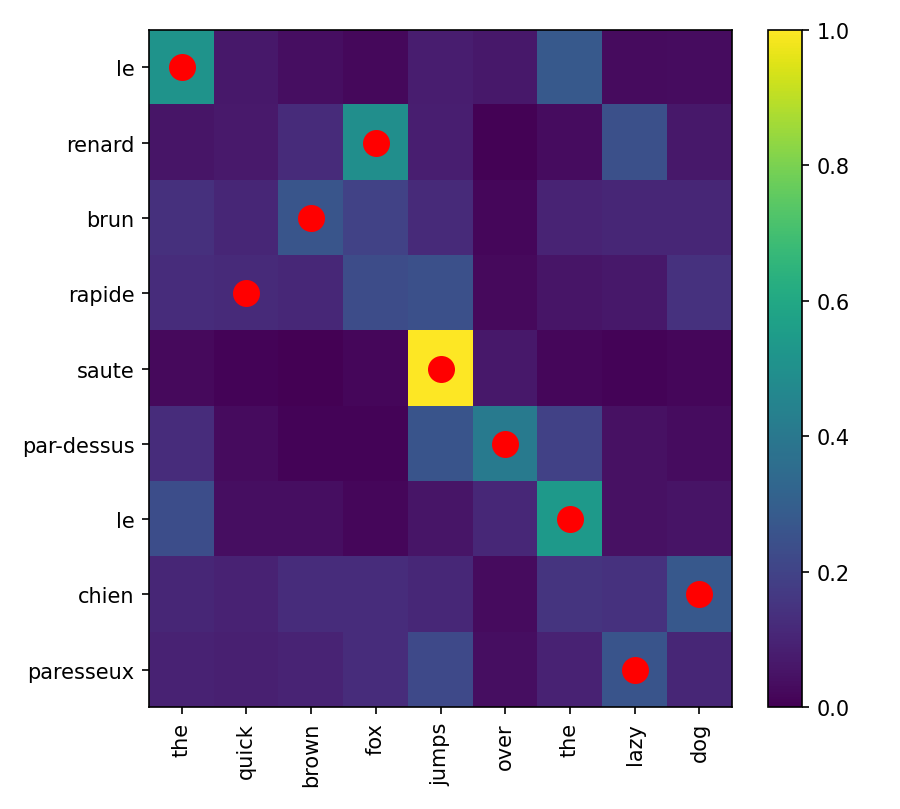}& \includegraphics[width=0.22\textwidth]{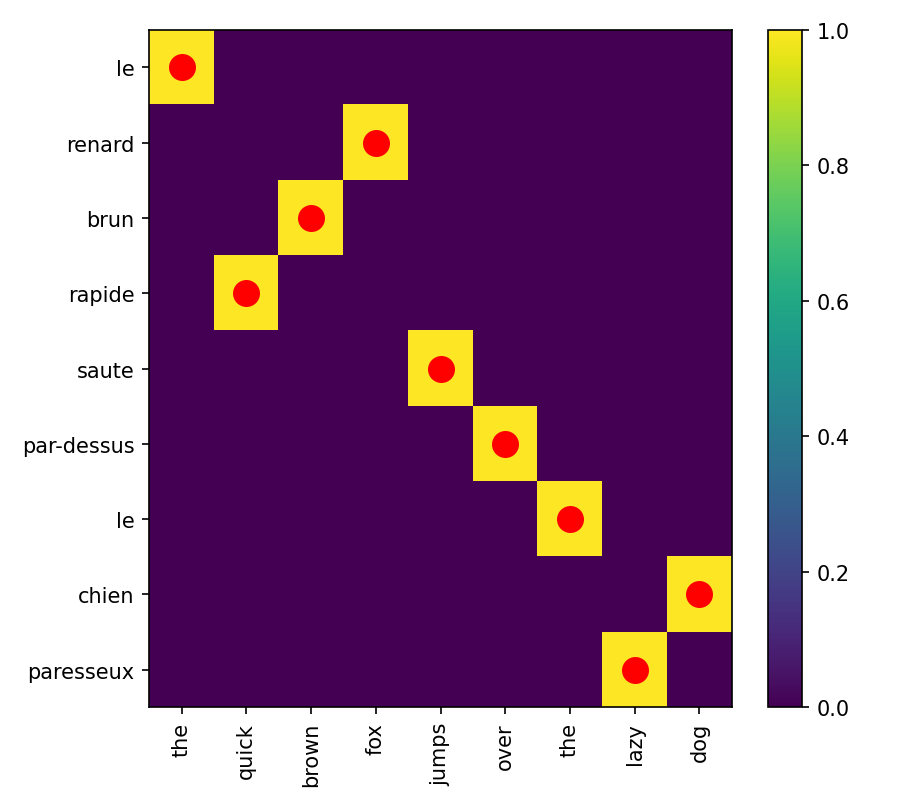}\\
    Attention at (1) & Attention at (6) & Attention at (12) & OT vs Translation
\end{tabular}
\caption{\footnotesize{\textbf{Translation with OT.} 
The rightmost plot shows the optimal transport (OT) solution from~\eqref{eq:optimal_transport}, computed between the English word embeddings ($x_1, \dots, x_n$) and the French word embeddings ($y_1, \dots, y_n$). Red dots mark correctly aligned translation pairs. 
Observe the OT solution matches words with the same meaning. 
The other plots depict attention-weight heatmaps from layers 1, 6, and 12, showing how the model iteratively approximates the OT solution.
}} \label{fig:translation}
\end{figure*}

Mechanistic interpretability of large language models has become a central research focus in machine learning, aiming to explain the internal mechanisms that underlie data generation in these models. Significant progress has been made in understanding the computational role of attention heads~\citep{elhage2021mathematical}, as well as the emergence of in-context learning capabilities across transformer layers~\citep{garg2022can}. Recent work on in-context learning reveals that the internal feature-extraction dynamics of transformers closely mirror first-order optimization methods~\citep{von2023transformers,akyurek2022learning,ahn2024transformers,lutz2025lineartransformersimplicitlydiscover}, providing both conceptual insight and theoretical guarantees for solving classical statistical inference problems such as linear regression with in-context learning. An important open question, however, is how these computational mechanisms relate to standard natural-language-processing tasks, including machine translation, text comprehension and sentiment analysis.   

This paper connects machine translation to the in-context learning of \emph{discrete optimal transport (OT)}. OT  has broad applications in data mining~\citep{peyre2019computational}, bio-informatics~\citep{schiebinger2019optimal}, and natural language processing~\citep{munkres1957algorithms}. Given two sets of $n$ points $ \{x_1,\dots,x_n\}$ and $ \{y_1,\dots,y_n\}$ in $\mathbb{R}^d$, the discrete OT problem seeks a permutation matrix $P^\ast \in \{0,1\}^{n\times n}$ that minimizes the total matching cost
\begin{equation}
P^\ast = \arg\min_{P\in\Pi_n} \sum_{i,j} P_{ij}\,\|y_i - x_j\|_2^2,\label{eq:optimal_transport}
\end{equation}
where $\Pi_n$ is the set of $n\times n$ permutation matrices. Sorting is a special case of discrete OT when $d=1$ and the $y_i$’s are ordered. We first show that solving the above problem relates to the mechanism of machine translation in LLMs.

\subsection{Translation with OT}
We connect optimal transport (OT) to the mechanism of language translation in LLMs by analyzing their internal word embeddings. As an illustrative example, we prompt \textsc{BERT} with the following pair of translated sentences in English and French:
\begin{equation*}
    \begin{cases}
        \text{\footnotesize{The quick brown fox jumps over the lazy dog}}\\
        \text{\footnotesize{Le renard brun rapide saute par-dessus le chien paresseux}}.
    \end{cases}
\end{equation*}

Let $x_1,\dots,x_n$ and $y_1,\dots,y_n$ denote the vector representations of the English and French words, respectively, from the first encoder layer. 
Fig.~\ref{fig:translation} shows that the OT solution  in~\eqref{eq:optimal_transport} aligns pairs of translated words.  We present further experiments on 10,000 sentences in Section \ref{section:OT_translation}, demonstrating that OT on word embeddings aligns words for translation.

Even more interesting is the pattern of average attention weights (over heads) across the transformer layers. Fig.~\ref{fig:translation} 
shows that the attention weights progressively approximate the OT solution ($P^*$ in~\eqref{eq:optimal_transport}) as depth increases. 
Remarkably, this interpretable mechanism emerges naturally after pretraining, without any explicit regularization. 
In other words, the transformer learns to align translated words across its layers. We present a more extensive experiment on 10,000 translated sentences on Marian model~\citep{junczys2018marian} in Section~\ref{section:experiments}, supporting the above observation.  
But how can we explain this striking algorithmic behavior of attention weights? To address this question, we perform additional experiments on a transformer specifically trained to solve OT.

\subsection{Observations on OT Mechanism}
\begin{figure*}[t!]
    \centering
    \begin{tabular}{ M{3.5cm}   M{2.5cm} M{2.5cm} M{2.5cm} |M{3cm}}

& test $n=9$& &  &  optimal $P^*$ 
\\
(1) generalization & \includegraphics[width=0.12\textwidth]{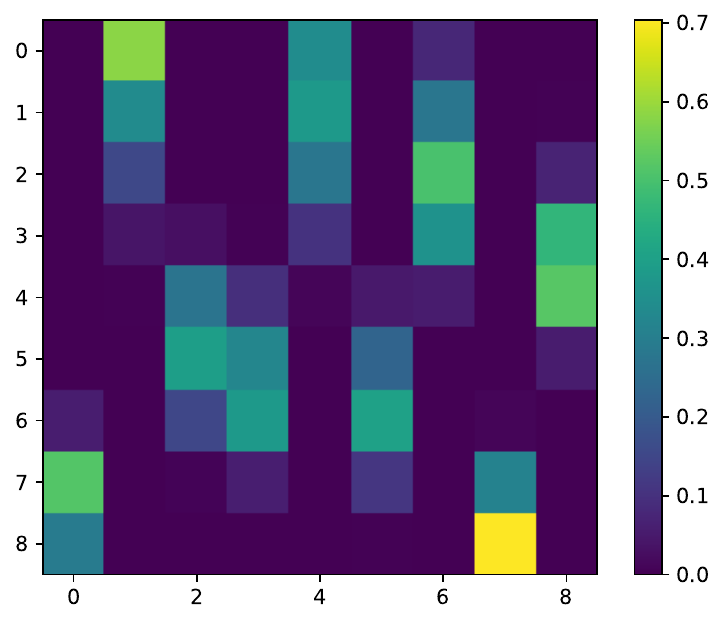} & &  & \includegraphics[width=0.12\textwidth]{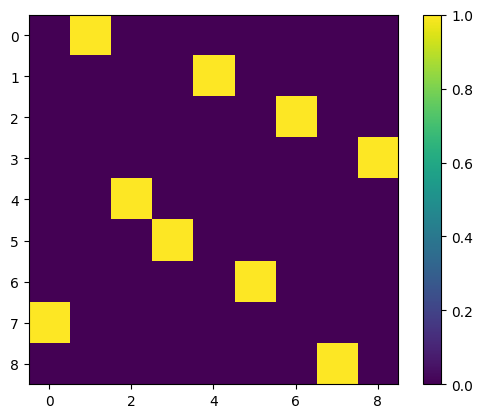}\\
&  
       without   & with 
& &   \\
(2) prompt engineering& \includegraphics[width=0.12\textwidth]{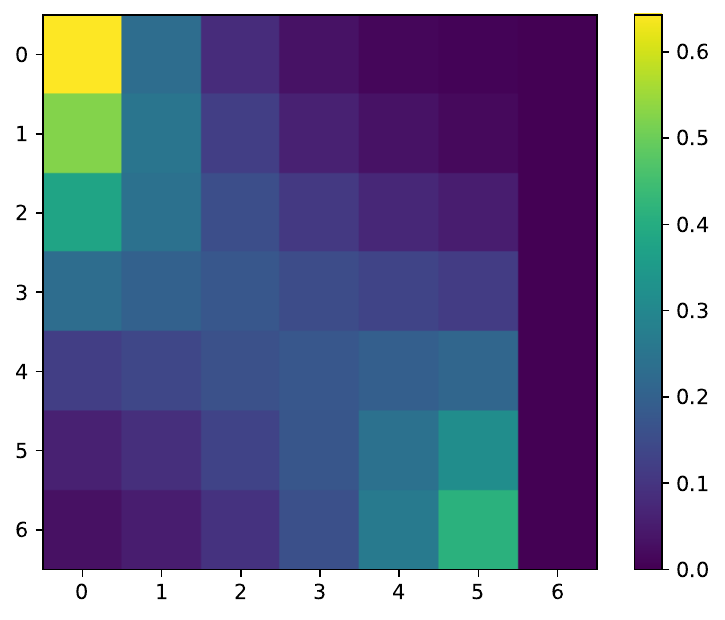}   & \includegraphics[width=0.12\textwidth]{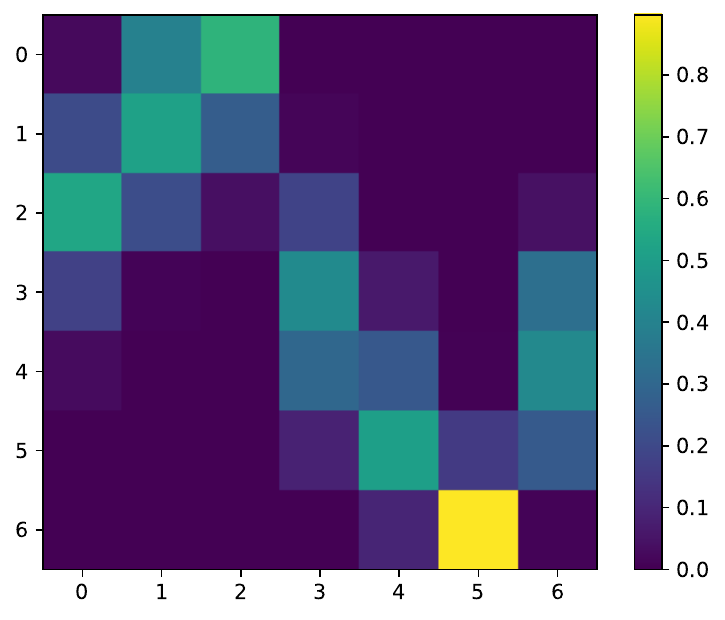}   & &\includegraphics[width=0.12\textwidth]{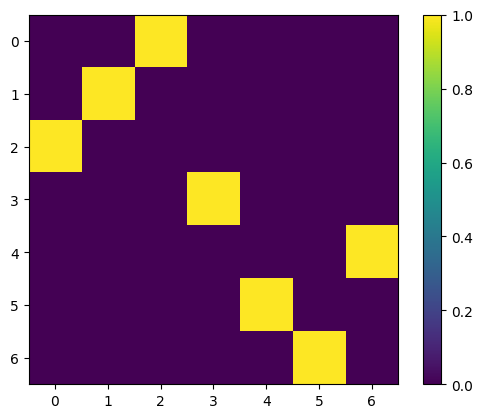} \\   

& layer 5 & layer 15 & last layer (20) &\\
(3) Iterative inference & \includegraphics[width=0.12\textwidth]{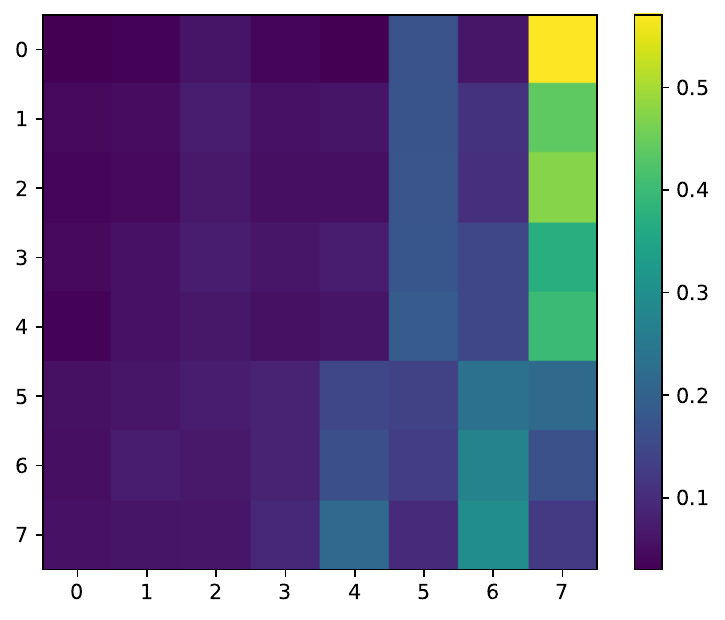}& \includegraphics[width=0.12\textwidth]{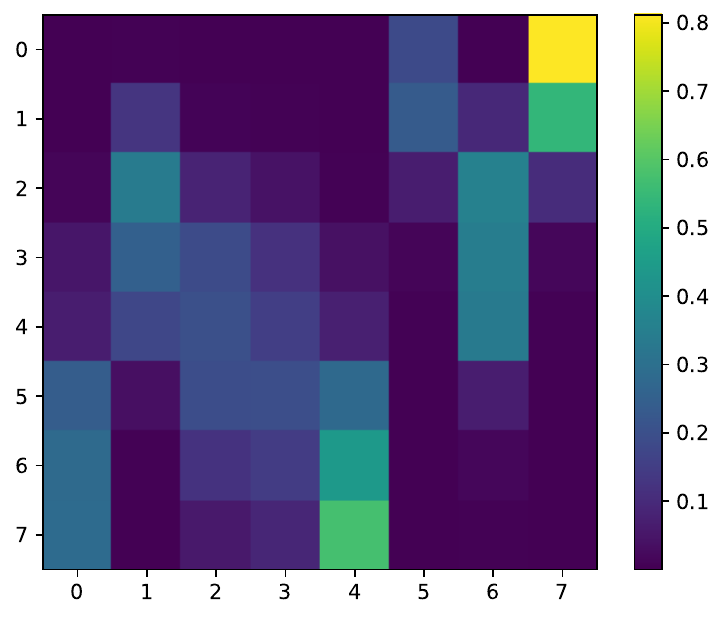} & \includegraphics[width=0.12\textwidth]{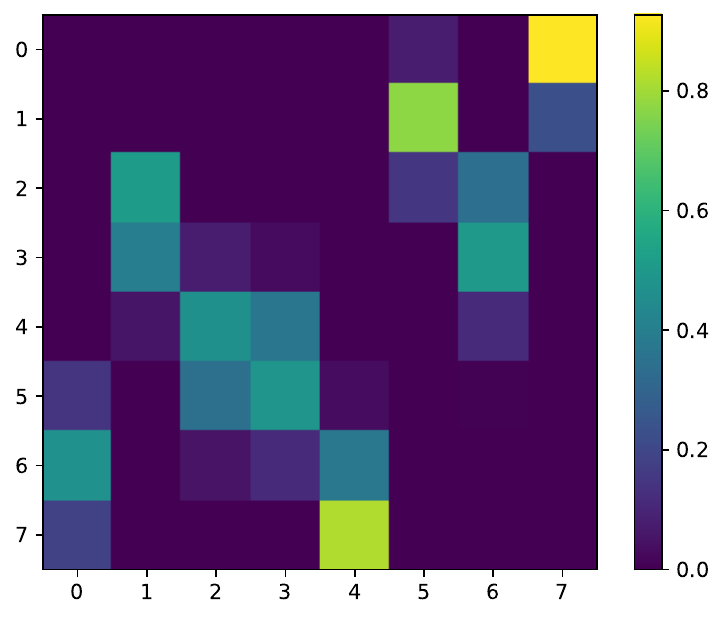}& \includegraphics[width=0.12\textwidth]{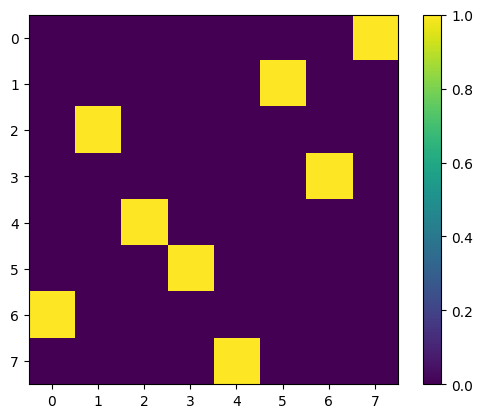} 
\end{tabular}\caption{\footnotesize{\textbf{Observations on In-context Learning for OT.}
\emph{(1)} The model is trained to solve OT with 7 data points and evaluated on 9 data points. 
The left image shows the attention weights, which closely approximate the OT solution shown on the right.
\emph{(2)} After specific prompt engineering, the attention weights between tokens estimate the OT solution. Notably, this prompt engineering is used in (1).
\emph{(3)} The attention weights evolve across layers, progressively yielding a more accurate approximation of the optimal solution. See Appendix~\ref{sec:experiments_app} for details.
}}
\label{fig:train}
\end{figure*}

To analyze the OT mechanism of attention, we trained a transformer to solve discrete OT on small problem instances with $n = 7$ points and evaluated it on larger instances with $n = 9$. For training, we generated synthetic data from a standard neutral distribution widely used to study in-context learning~\citep{garg2022can} (see Appendix~\ref{sec:experiments_app} for details). Surprisingly, the transformer can approximate the optimal OT solutions on larger inputs (Fig.~\ref{fig:train}). This out-of-distribution generalization suggests that transformers can adapt to input sizes beyond those seen in training, a key requirement for in-context learning. This result is particularly important given prior efforts to modify transformers for solving OT~\cite{sander2022sinkformers,tay2020sparse}. Yet, Fig.~\ref{fig:train} demonstrates that standard transformers can already handle sorting and, more broadly, the optimal transport problem.

A critical factor behind the observed generalization is \emph{prompt engineering}. Careful input augmentation effectively extends the transformer’s memory, thereby significantly enhancing its computational capacity to solve OT; see Observation~(2) in Fig.~\ref{fig:train}. We provide further details and insights on this specific prompt construction in the following sections.

The proposed prompt design encourages the attention layers to solve the OT problem iteratively as depth increases. As shown in Fig.~\ref{fig:train}, initially diffuse attention weights gradually converge layer by layer toward the optimal OT solution. While such \emph{iterative inference} has been analyzed for linear regression~\citep{ahn2024transformers,lutz2025lineartransformersimplicitlydiscover}, it remains unclear whether these results extend beyond linear regression.

\subsection{Contributions}
We establish three main contributions:
\paragraph{Contribution~(1): Mechanistic Analysis.}
We discover a link between the mechanism of transformer for language translation and OT. Our observations show that attention weights align translated words iteratively across the layers, which approximates OT solution for word embeddings.   

To explain this striking observation, we analyze the internal dynamics of feature extraction in transformers. We show that the fundamental building block softmax self-attention~\citep{dong2021attention} is particularly well suited to \emph{implement} OT. In particular, a single self-attention layer can simulate an iterative first-order optimization method for OT, thereby explaining the iterative inference behavior observed in Fig.~\ref{fig:train}. 

\medskip
\paragraph{Contribution~(2): Theoretical guarantees.} Leveraging the mechanistic insight, we prove: for any two sets of $n$ points in $\mathbb{R}^d$, a transformer can estimate the OT solution with
\begin{align}
O\!\left(\tfrac{n^{3/2}}{\mathrm{depth}^{1/2}}\right)\text{-accuracy for all integers } n.
\end{align}
Consequently, sufficiently deep transformers can solve OT for a wide range of $n$ without modifying their parameters, explaining scale generalization, Observation~(1) in Fig.~\ref{fig:train}.

\medskip
\paragraph{Contribution~(3): Insights on prompt engineering.} Crucially, the established results hinges on \emph{prompt engineering}, which provides an extended memory for the attention layers. It also offers a new insights  the mechanism of prompt engineering.



\section{Background}
\subsection{Entropy regularization}
While OT is constrained to the combinatorial set of permutation matrices, it can be relaxed to optimization over a continuous set.  The state-of-the-art method is based on linear programming, namely
\begin{align*}
  \widehat{P} &= \arg\min_{P \in \mathbb{R}^{n\times n}} \sum_{ij} P_{ij} C_{ij}, \\ 
  & \text{ subject to } P \text{ is doubly stochastic},
\end{align*}
where $C \in \R^{n\times n}$ is a cost matrix such that $C_{ij}=\|y_i - x_j\|^2$.
Comparing the above problem with the original problem in \eqref{eq:optimal_transport}, we notice that the constraint requiring $P$ to be a permutation matrix  has been relaxed to allowing $P$ to be a doubly stochastic matrix. Recall the solutions of linear programs lie among the extreme points of the constraint set. Since the extreme points of doubly stochastic matrices are permutation matrices~\citep{conte1991gradient}, the above linear program has the same solution as \eqref{eq:optimal_transport}, i.e., $P^* = \widehat{P}$.

The above linear program has $O(n^2)$ variables. Due to the quadratic growth with $n$, solving the linear program becomes computationally challenging for large $n$. \cite{cuturi2013sinkhorn} proposes a computationally efficient alternative  based on regularization with entropy: 
\begin{equation}
\begin{aligned}
 \label{eq:ot_reg}
     P^*_\lambda &:= \arg\min_{P \in \R^{n\times n}} \sum_{ij} P_{ij} C_{ij} + \lambda  P_{ij} \log(P_{ij}), \\
     &\text{subject to $P$ is doubly stochastic}
\end{aligned}
\end{equation}
The Lagrangian dual of the above program has only $O(n)$ variables which is considerably fewer than $O(n^2)$ variables for the original linear program. Introducing the dual parameters $v \in \R^n$ and $u \in \R^n$, the Lagrangian function is defined as follows: 
\begin{multline*}
     \sum_{ij} P_{ij} C_{ij} + \lambda \sum_{ij} P_{ij} \log(P_{ij})  \\ - u^\top \left(P 1_n -1_n\right) - v^\top \left(P^\top 1_n - 1_n\right).
\end{multline*}
$P_{ij} = e^{\frac{-C_{ij}+v_j +u_i}{\lambda}-1}$ minimizes the Lagrangian function. This structure inspired the use of Sinkhorn’s fixed point iteration to find the solution of the dual problem.  In particular,  \cite{sinkhorn1967diagonal} proves that there exists a unique doubly stochastic matrix of the form $[P_{\lambda}^*]_{ij} = e^{\frac{-C_{ij}+v_i^* +u_j^*}{\lambda}-1}$ that is the solution of a simple fixed-point iteration where $u^*, v^*$ are unique up to scaling factors. $[M]_{ij}$ denotes the element of the matrix $M$ located in row $i$ and column $j$. Leveraging this theorem, \cite{cuturi2013sinkhorn} proves Sinkhorn recurrence can efficiently find $P^*_\lambda$.

Apart from  Sinkhorn’s recurrence other optimization method can also  solve Lagrangian dual problem. Recall the minimizer $P_{ij} = e^{\frac{-C_{ij}+u_i +v_j}{\lambda}-1}$. Plugging this into the Lagrangian function yields
\begin{align} \label{eq:L}
  \min_{v, u \in \R^n}     \underbrace{\lambda \left( \sum_{ij} e^{\tfrac{-C_{ij} + u_i +v_j}{\lambda}-1}  \right)  - \sum v_i - \sum_{i} u_i}_{L(u,v)} 
\end{align} 

It is easy to check that $L$ is convex in $u$ and $v$ as its Hessian is diagonally dominant, hence positive semi-definite. Thus, standard first-order optimization can optimize $L$ such as gradient descent with adaptive coordinate-wise stepsizes:  \begin{align} \label{eq:gd}
\begin{cases}
    u_{\ell+1} = u_{\ell} -  D_{\ell} \nabla_u L(u_{\ell},v_{\ell}) \\ 
    v_{\ell+1} = v_{\ell} - D_{\ell}' \nabla_v L(u_{\ell},v_{\ell})
\end{cases},  
\end{align}
where $\nabla_u L$ denotes the gradient of $L$ with respect to $u$ and $D_\ell, D_\ell' \in \R^{n\times n}$ are diagonal matrices with positive diagonal elements which includes stepsize for each coordinate of vectors $u_\ell$ and $v_\ell$. We will prove that self-attention in transformers can simulate the above iterations. 

\subsection{Soft-max Self-attention}
Attention layers are fundamental building blocks of neural networks, developed over decades of research. \cite{hochreiter1997long} pioneered this development by proposing an attention mechanism for Recurrent Neural Networks (RNNs) inspired by human cognition. \citet{graves2014neural} employs the attention mechanism to develop a memory system for a parametric version of the Turing machine. \citet{bahdanau2014neural} adapts this attention mechanism in neural Turing machine to design a powerful model for machine translation. While attention was originally introduced for recurrent models, \citet{vaswani2017attention} proposes non-recurrent attention layers, combined with residual connections~\citep{he2016deep}, thereby significantly enhancing the training of attention weights.  

Attention layers rely on a convex combination. Let $Z \in \R^{n\times d}$. An attention layer is a function denoted by  $\attn_{\theta}: \R^{n\times d} \to \R^{n\times d}$ with parameters $\theta := \left[K, Q, V \in \R^{d \times d} \right]$ defined as 
\begin{align*}
    \attn_{\theta}(Z) =   A Z V, \quad A_{ij} = \tfrac{e^{\langle K z_i, Q z_j \rangle}}{\sum_{j=1}^n e^{\langle K z_i, Q z_j \rangle}},  
\end{align*}
where $z_i$ and $z_j$ are rows of $Z$.
The convex combination of data points induces a local dependency between the representations of tokens, thereby capturing the contextual relationships among them.

\citet{tay2020sparse} investigate whether attention layers can perform sorting. 
Since standard self-attention layers cannot directly implement Sinkhorn’s iteration, 
\citet{tay2020sparse,sander2022sinkformers} propose a novel attention mechanism called \emph{Sinkhorn attention}. 
However, because standard self-attention remains widely used, we investigate whether the standard self-attention can approximate the OT solution.

\subsection{The Engineered Prompt}
 We propose a particular input denoted by $Z_{0} \in \R^{(n+1) \times (2d+9)}$ to encode the assignment problem:
\begin{equation} \label{eq:engineered_prompt} 
    Z_0 = \begin{bmatrix}
    \textcolor{blue}{x_1} &  \textcolor{blue}{y_1} & \textcolor{red}{\|x_1\|^2} & \textcolor{red}{\|y_1\|^2} &  \textcolor{red}{1_4} & \textcolor{red}{0_3} \\ 
    \textcolor{blue}{x_2} &  \textcolor{blue}{y_2} & \textcolor{red}{\|x_2\|^2} & \textcolor{red}{\|y_2\|^2} & \textcolor{red}{1_4}&  \textcolor{red}{0_3}\\
    \textcolor{blue}{\vdots} & \textcolor{blue}{\vdots} & \textcolor{red}{\vdots}& \textcolor{red}{\vdots} & \textcolor{red}{\vdots}   \\
    \textcolor{blue}{x_n} &  \textcolor{blue}{y_n} & \textcolor{red}{\|x_n\|^2} & \textcolor{red}{\|y_n\|^2} & \textcolor{red}{1_4} & \textcolor{red}{0_3}  \\
      \textcolor{blue}{0_d} & \textcolor{blue}{0_d} & \textcolor{red}{0} & \textcolor{red}{0} &  \textcolor{red}{-v_4}& \textcolor{red}{0_3}  
    \end{bmatrix}.
\end{equation}

The elements in \textcolor{blue}{blue} are the original prompts, which are sufficient for the assignment. The elements in \textcolor{red}{red} are carefully engineered. $1_4$ denotes the all-ones 4-dimensional vector, $0_d$ denotes the all-zeros d-dimensional vector, and $v_4 = [0, 0, 0, 1]$.

Our analysis shows that the engineered prompt gives the transformer enough input structure and memory to simulate optimization methods for optimal transport (OT). The appended zeros create an augmented representation across layers, allowing attention to store and update gradient-descent iterates. In experiments, using the natural encoding of \((x_i, y_i)\) in the prompt reduces performance (see Observation~(2) in Fig.~\ref{fig:train}).

\subsection{Transformer}
We consider a specific transformer architecture composed of multiple soft-max self-attention with residual connections. Let \(Z_\ell\) denote the intermediate representation of the input \(Z_0\) at layer \(\ell\) which obeys the following recurrence
\begin{equation} \label{eq:Zdynamics} Z_{\ell+1} = Z_{\ell}+ \sum_{j=1}^2 \attn_{\theta_j^{(\ell)}}(Z_{\ell}) B_j^{(\ell)}, 
\end{equation}
where \(\theta_j^{(\ell)}\) are parameters of the attention layers, \(B_j^{(\ell)} \in \mathbb{R}^{d' \times d'}\) are the weights to combine attention heads. Remarkably, the model has two attention heads. 
\begin{figure*}[t!]
    \centering
    \begin{tabular}{c  c c |c}
\includegraphics[width=0.18\textwidth]{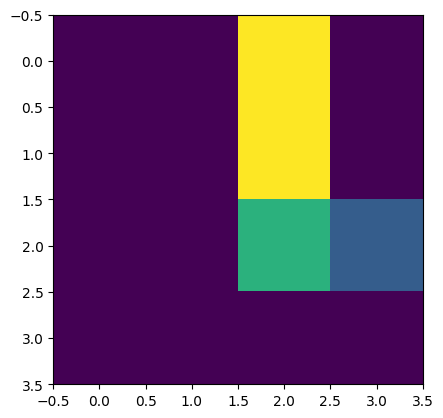}  & \includegraphics[width=0.18\textwidth]{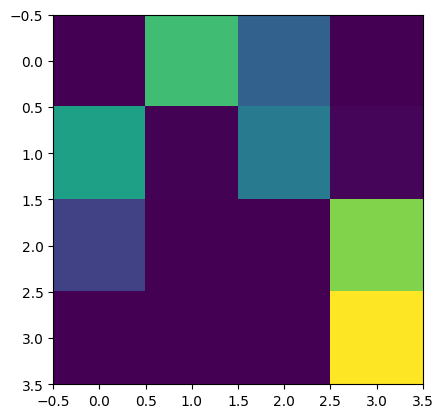}  &  \includegraphics[width=0.18\textwidth]{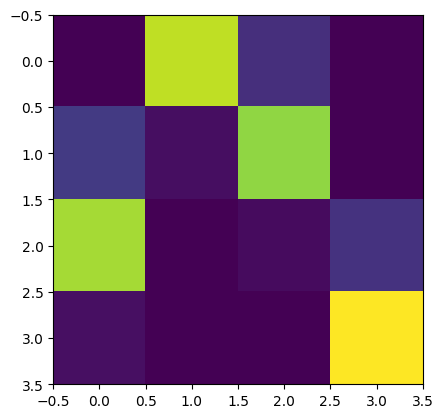}  & \includegraphics[width=0.2\textwidth]{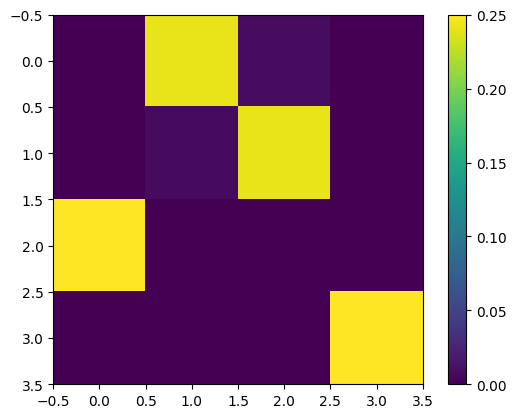} \\
     $A^{(1)}$    & $A^{(300)}$ & $A^{(600)}$ & $P^*_\lambda$
    \end{tabular}
    \caption{\footnotesize{\textbf{Convergence of attention matrices.}  The plotted matrices are attention weights in layers (1), (300) and (600). We observe these matrices converge to the regularized OT solution (the rightmost plot), which is proven by  Theorem~\ref{thm:convergence}. }}
    \label{fig:convergence}
\end{figure*}
Remarkably, all our results extend to transformer blocks that include both attention and feedforward layers. 
This is because feedforward layers with zero weights, combined with residual connections, effectively reduce the architecture to one consisting solely of self-attention layers. 
Moreover, restricting the analysis to two attention heads is not a limitation: any transformer with more heads can be reduced to an equivalent two-head attention block by zeroing out the mixing weights of the additional heads. 
Finally, although we do not explicitly include attention masks in our formulations, appropriate masking can be incorporated without affecting the validity of our analysis.

\subsection{Notations}
 \begin{table}[h!]
\centering
\caption{Notation summary}
\label{tab:notation}
\begin{tabular}{ll}
\toprule
\textbf{Symbol} & \textbf{Definition} \\
\midrule
$n$ & The number of points in OT\\
$d$ & Dimension\\
$\{x_i/y_i\}_{i=1}^n$ & Sets of points in $\mathbb{R}^d$ for OT \\
$C \in \mathbb{R}^{n \times n}$ & Cost with  $C_{ij} = \|y_i - x_j\|_2^2$ \\
$\Pi_n$ & Set of permutation matrices \\
$P^{*} \in \{0,1\}^{n\times n}$ & Optimal transport in \eqref{eq:optimal_transport} \\
$\lambda > 0$ & Regularization parameter in \eqref{eq:ot_reg} \\
$P^{*}_{\lambda} \in \mathbb{R}^{n\times n}$ & Optimal regularized OT plan\\
$u, v \in \mathbb{R}^n$ & Dual parameters \\
$Z_\ell \in \mathbb{R}^{n\times d}$ & Token representations at layer $\ell$ \\
$A^{(\ell)} \in \mathbb{R}^{n\times n}$ & Attention weight at layer $\ell$ \\
\bottomrule
\end{tabular}
\end{table}
Let \( Z_0 \in \mathbb{R}^{n \times d} \) denote the input to the transformer. Its representation at layer \( \ell \) is denoted by \( Z_\ell \). The input matrix \( Z_0 \) contains the points \( x_1, \dots, x_n \) and \( y_1, \dots, y_n \) from the assignment problem. Let \( P^* \) be the optimal assignment solution, and let \( P^*_\lambda \) be its approximation obtained via entropy regularization. The transformer is parameterized by \( \theta \), which includes all attention parameters \( \theta^{(\ell)}_j \) at layer \( \ell \) and attention head \( j \).
The vectors \( u_\ell \) and \( v_\ell \) denote the gradient descent iterates on the dual objective function \( L \), as defined in~\eqref{eq:L}. For a matrix \( M \), we use the notation \( [M]_{i,j} \) to refer to the entry in row \( i \) and column \( j \). See Table~\ref{tab:notation} for summary.

\section{The  Mechanism of Softmax Attention for OT} \label{sec:convergence}
What is given to a language model is ultimately a vector representation of words (word embeddings). How does language model matches words with the same meaning given the word embeddings? 
 Fig.~\ref{fig:train} shows the process of alignment is iterative across the layers. We mathematically prove this iterative alignment by linking attention weights to dual OT. In particular, the next theorem states that $\ell$ attention layers stacked in transformers can simulate $\ell$-iterations of gradient descent on the dual objective function $L$ defined in \eqref{eq:L}.

\begin{boxtheorem}[Dual OT with Transformers]\label{thm:gd}
 Let $D_\ell, D'_\ell \in \R^{n\times n}$ are diagonal matrices whose diagonal elements are
  \begin{align*}
  \begin{cases}
       (\gamma_\ell [D_{\ell}]_{ii})^{-1} =  \sum_{j} e^{(-C_{ij}+[u_\ell]_i+[v_\ell]_{j})/\lambda-1}+1,\\   (\gamma_\ell [D_{\ell}']_{jj})^{-1} =  \sum_{i} e^{(-C_{ij}+[u_\ell]_i +[v_\ell]_j)/\lambda-1}+1.
  \end{cases}
\end{align*}
   There exists a configuration of parameters \emph{independent from inputs and $n$} such that 
    \begin{align*}
      \begin{cases}[Z_{\ell}]_{(1:n),(2d+7)} = u_{\ell}-D_\ell \nabla_u L(u_{\ell},v_{\ell}) \\
      [Z_{\ell}]_{(1:n),(2d+8)} = v_{\ell}-D_\ell' \nabla_v L(u_{\ell},v_{\ell})
      \end{cases},  
    \end{align*}
    holds for all integer values of $n$, where $u_{\ell}$ and $v_{\ell}$ are gradient descent in ~\eqref{eq:gd} 
 iterations starting from $u_0 = v_0 = 0$.  
\end{boxtheorem}

Notably, the above result supports the \emph{"iterative inference hypothesis"} \citep{jastrzkebski2017residual}, that links the mechanism of deep neural networks to optimization methods. This hypothesis postulates that residual connections enable deep networks to implicitly implement gradient descent across layers to tackle complex tasks. The hypothesis is based on striking observations on the underlying mechanisms of Convolutional Neural Networks (CNNs) \citep{alain2016understanding}. Previous studies have theoretically proven this hypothesis for solving least-squares using transformers~\citep{ahn2024transformers,von2023transformers,akyurek2022learning}. Building on these studies, we demonstrate that transformers can implement gradient descent for OT and connect it to language translation.

We highlight that the established result holds for standard \textbf{softmax self-attention}. In particular, we show that softmax attention is especially well-suited to implementing gradient descent on the objective function $L$, as defined in~\eqref{eq:L}. Remarkably, attention mechanisms cannot simulate the gradient of an arbitrary function and exhibit inherent limitations in their computational capabilities. In particular, \emph{linear attention} is well-suited for problems such as linear regression, where the underlying objective is quadratic~\citep{ahn2024transformers,von2023transformers}. In contrast, \emph{softmax attention} is better aligned with tasks involving token matching.

Prompt engineering is essential for the proof of Theorem~\ref{thm:gd}. Expanding the input size by adding columns and rows creates an extended data representation matrix across the layers. Attention layers can utilize a part of this matrix \emph{as memory to store the iterates of gradient descent} . Furthermore, the input dependent part of the prompt supplies the necessary statistics for the attention layers to implement gradient descent.

By connecting the intermediate data representations to a well-established algorithm, we can leverage powerful theoretical tools to prove that a transformer with frozen weights can approximate OT solution. We first present an informal version of the result below and defer the full theorem and proof to Appendix \ref{app:convergence}.

\begin{figure*}[t!]
    \centering
    \begin{tabular}{c c | c c}
\includegraphics[width=0.18\textwidth]{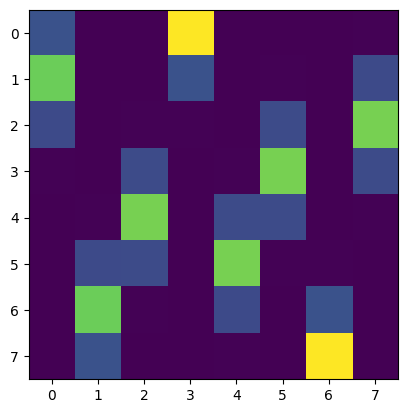}   & \includegraphics[width=0.18\textwidth]{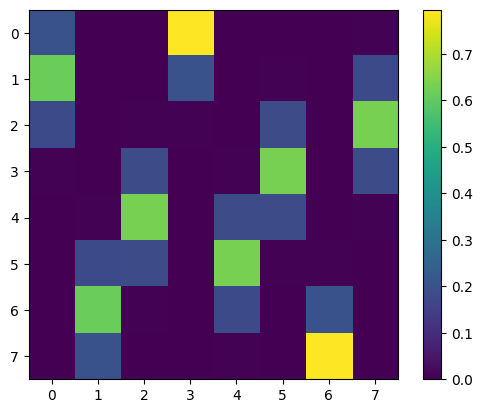} 
         & \includegraphics[width=0.18\textwidth]{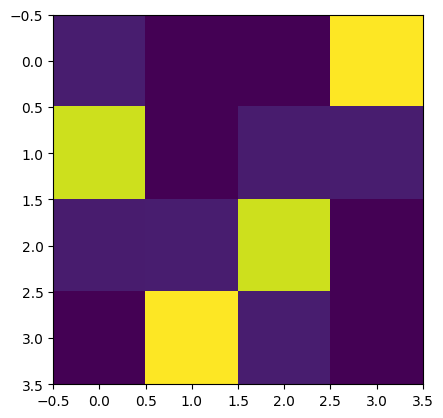}   & \includegraphics[width=0.18\textwidth]{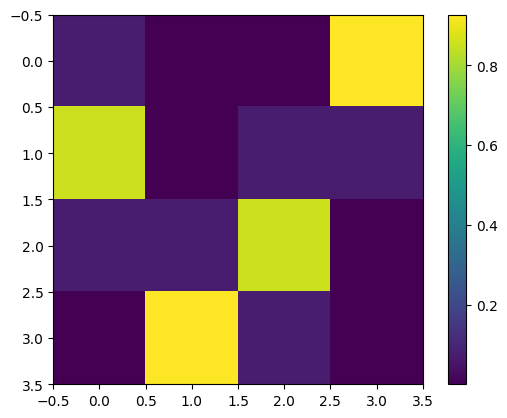} \\
         $A^{(2000)}$ & $P^*_\lambda$ & $A^{(2000)}$ & $P^*_\lambda$
    \end{tabular}
\caption{\footnotesize{\textbf{Sample Size.} left: $n=8$, right: $n=4$. The transformer's weights remain unchanged. Observe the transformer can solve the OT problem for both values of \(n\), demonstrating a form of out-of-distribution generalization proven in Thm.~\ref{thm:gd}.}}
    \label{fig:n}
\end{figure*}

\begin{boxtheorem}[Approximation Bound] \label{thm:convergence}
   \textit{Informal statement:} There exists a choice of attention parameters, independent of $n$, $d$, and the input $Z_{0}$, such that the attention weights at layer~$\ell$ approximate the regularized optimal transport solution $P_{\lambda}^{*}$ with error
\[
O\!\left(\frac{n^{3/2} e^{1/\lambda}}{\sqrt{\ell}}\right),
\]
where $\lambda>0$ is the regularization parameter.

\end{boxtheorem}
The error bound provided above is in terms of  the Hilbert projective metric, a well-established measure for analyzing the convergence behavior of the Sinkhorn algorithm \cite{franklin1989scaling}. According to the theorem, the attention matrices converge to $P^*_\lambda$ at a rate of $O\left(\frac{1}{\text{depth}^{\sfrac{1}{2}}}\right)$, implying that performance improves with increasing depth. This convergence holds for any integer \( n \). Thus, a sufficiently deep transformer can approximate solution of OT for broad choice of $n$ without changes in parameters. This result mathematically proves transformers are capable of seen generalization in Fig.~\ref{fig:train}.

An application of the last theorem is that transformers can be used to sort lists. As discussed, sorting is a specific instance of assignment problem for \(d=1\) with \(y_1 < \dots < y_n\). Thus, transformer can sort with an error that vanishes with $\lambda$. \citet{graves2014neural} experimentally demonstrate that the neural Turing machine can sort. Here, we theoretically prove this capability for transformers by establishing an approximation bound.
 
Notably, the proof of the last theorem is constructive; it provides an explicit form of the parameters. 
This explicit characterization enables us to experimentally validate the result by directly instantiating the parameter choices. 
Fig.~\ref{fig:convergence} shows that the estimate of the OT solution improves as the network depth increases, confirming the inverse dependence of the bound in the theorem on depth. 
Similarly, Fig.~\ref{fig:n} demonstrates that a transformer with frozen weights can solve OT for varying values of $n$, explaining the out-of-distribution  generalization observed in Fig.~\ref{fig:train}.
 
\section{Experiments}
So far, we have provided a mechanistic analysis of token alignment in LLMs, showing that attention weights can iteratively align tokens across transformer layers. In this section, we investigate whether such token alignment also emerges in language translation.  
\subsection{Attention Alignment for Translation} \label{section:experiments}

According to Theorem~\ref{thm:convergence}, the attention weights should progressively improve token alignment across layers. 
To examine this phenomenon in a practical machine translation setting, we analyze pretrained Marian models for multilingual translation~\citep{junczys2018marian}. 

We sample $10{,}000$ English–French sentence pairs from the \textsc{WMT14} dataset~\citep{wmt}. 
All sentences are tokenized using the model’s built-in tokenizer. 
Because a tokenizer may split a single word into multiple subword tokens, we reconstruct word-level representations by averaging the final hidden states of the subword tokens composing each word.
To obtain reliable ground-truth word correspondences, we use the MUSE English–French dictionary~\citep{lample2017unsupervised}, which provides semantically equivalent translation pairs of words.

For each transformer layer $\ell$, the model outputs per-head self-attention matrices $\{A^{(\ell,k)}\}_{k=1}^H$. 
We average across heads to obtain
\[
A^{(\ell)}=\frac{1}{H}\sum_{k=1}^H A^{(\ell,k)} \in \mathbb{R}^{n\times n},
\]
and then extract the French$\rightarrow$English block when the inputs consist of concatenated English–French sentence pairs.
We use these matrices to assess how well the attention mechanism aligns translated words.

We quantify alignment quality with two complementary measures: \emph{Mean Reciprocal Rank (MRR)} and \emph{Hits@}$k$. 
Let $\mathcal{G}=\{(e_i,f_i)\}_{i=1}^n$ denote the set of English–French word pairs in a sentence, and let $A^{(\ell)}\in\mathbb{R}^{n\times n}$ denote the cross-attention matrix at decoder layer $\ell$, where rows correspond to French target words and columns to English source words. 
For each pair $(e_i,f_i)$ we define its \emph{rank} at layer $\ell$ as
\[
\mathrm{rank}^{(\ell)}(e_i,f_i)
=1+\bigl|\{\,e' : A^{(\ell)}_{f_i,e'}
      > A^{(\ell)}_{f_i,e_i}\,\}\bigr|,
\]
that is, one plus the number of English words receiving strictly higher attention from $f_i$ than the correct source $e_i$. 
The \emph{Mean Reciprocal Rank} (MRR) at layer $\ell$ is
\begin{equation} \label{eq:mrr}
\mathrm{MRR}^{(\ell)} = \frac{1}{|\mathcal{G}|}
\sum_{(e_i,f_i)\in\mathcal{G}}
\frac{1}{\mathrm{rank}^{(\ell)}(e_i,f_i)},
\end{equation}
while \emph{Hits@}$k$ at layer $\ell$ is the fraction of word pairs whose correct match is among the top-$k$ attention targets:
\begin{equation} \label{eq:hits}
\mathrm{Hits@}k^{(\ell)} =
\frac{1}{|\mathcal{G}|}
\sum_{(e_i,f_i)\in\mathcal{G}}
\mathbf{1}\!\bigl\{\mathrm{rank}^{(\ell)}(e_i,f_i)\leq k\bigr\}.
\end{equation}

$\mathrm{MRR}^{(\ell)}$ captures the average inverse rank of the correct translation, whereas $\mathrm{Hits@}k^{(\ell)}$ reports the proportion of pairs whose correct translation lies within the top-$k$ attended words.

\begin{figure}[h!]
    \centering
    \includegraphics[width=0.6\linewidth]{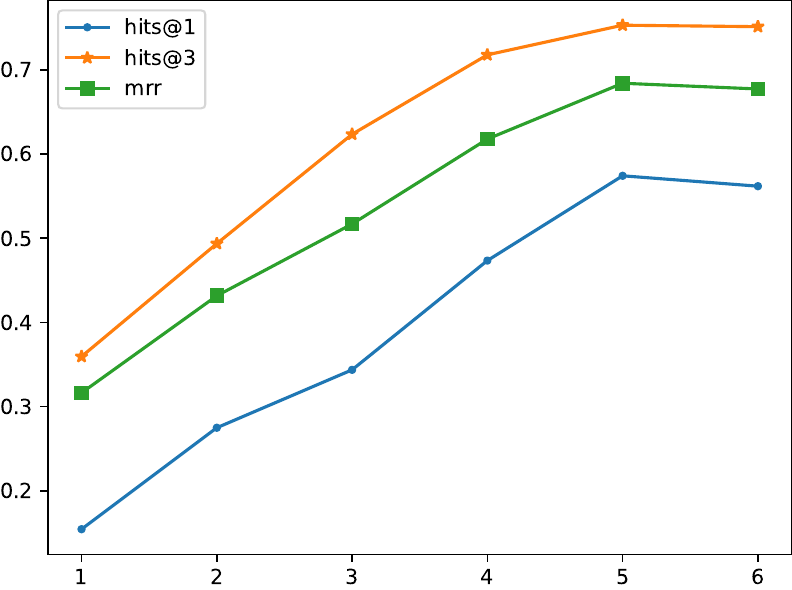}
    \caption{\footnotesize{\textbf{Attention dynamics for translation (En-Fr).} 
x-axis: transformer layer index~$\ell$. 
y-axis: The metrics defined in \eqref{eq:mrr} and \eqref{eq:hits}; Increasing metric values indicate that attention weights across layers progressively provide better estimates of translated word alignments, closely resembling the attention dynamics for OT.}}
    \label{fig:attention_algiment}
\end{figure}

Fig.~\ref{fig:attention_algiment} shows that both metrics consistently improve with depth: deeper layers produce attention maps that more accurately align translated word pairs. 
This iterative refinement parallels Theorem~\ref{thm:convergence}, confirming that attention in deeper layers converges toward the optimal transport solution.

\begin{figure}[h!]
    \centering
    \includegraphics[width=0.6\linewidth]{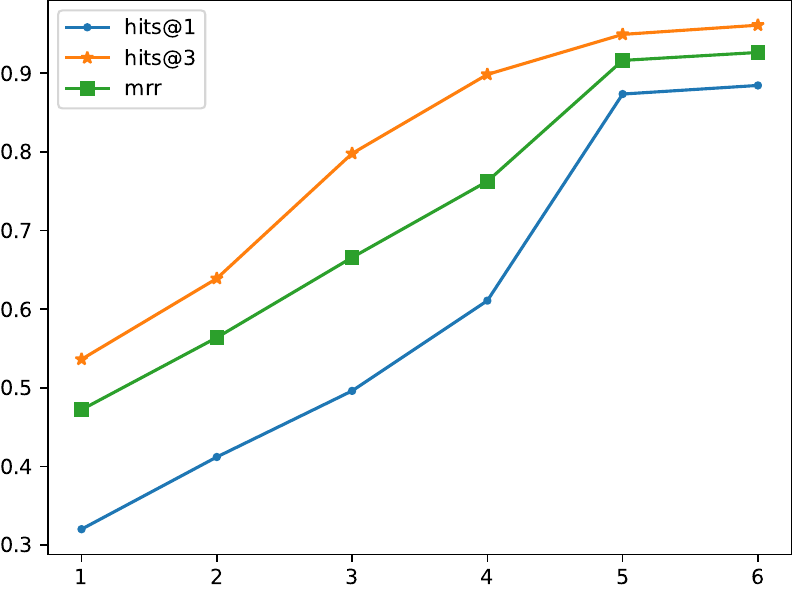}
    \caption{\footnotesize{\textbf{Attention dynamics for translation (DE-EN).} 
x-axis: transformer layer index~$\ell$. 
y-axis: The metrics defined in \eqref{eq:mrr} and \eqref{eq:hits}; Increasing metric values indicate that attention weights across layers progressively provide better estimates of translated word alignments, closely resembling the attention dynamics for OT.}}
    \label{fig:attention_algiment_de}
\end{figure}

This observation extends to other languages. In particular, we repeated the same experiment for German--English translation and found even stronger alignment of attention weights on translated word pairs. As shown in Figure~\ref{fig:attention_algiment_de}, attention in the final layer aligns over 80\% of the translated pairs.
   
\subsection{Embedding OT and Translation} \label{section:OT_translation}

We experimentally link the alignment of translated words to OT, substantiating the observation in Fig.~\ref{fig:train} using a large-scale evaluation on $10{,}000$ sentence pairs from the WMT14 English–French translation dataset~\citep{wmt}. 
Sentences are filtered to have between $3$ and $60$ tokens in each language after basic preprocessing (lowercasing and folding Unicode punctuation into ASCII).

To obtain ground-truth word alignments, we use the MUSE bilingual dictionary~\citep{lample2018word,conneau2018word}, which provides type-level translation pairs. 
We also use MUSE’s aligned monolingual word vectors for English and French (\texttt{wiki.multi.en.vec}, \texttt{wiki.multi.fr.vec}) to construct word embeddings. 
Because OT relies on inner products, we normalize all word embeddings to unit norm before computing the transport cost.

\begin{figure}[h!]
    \centering
    \includegraphics[width=0.7\linewidth]{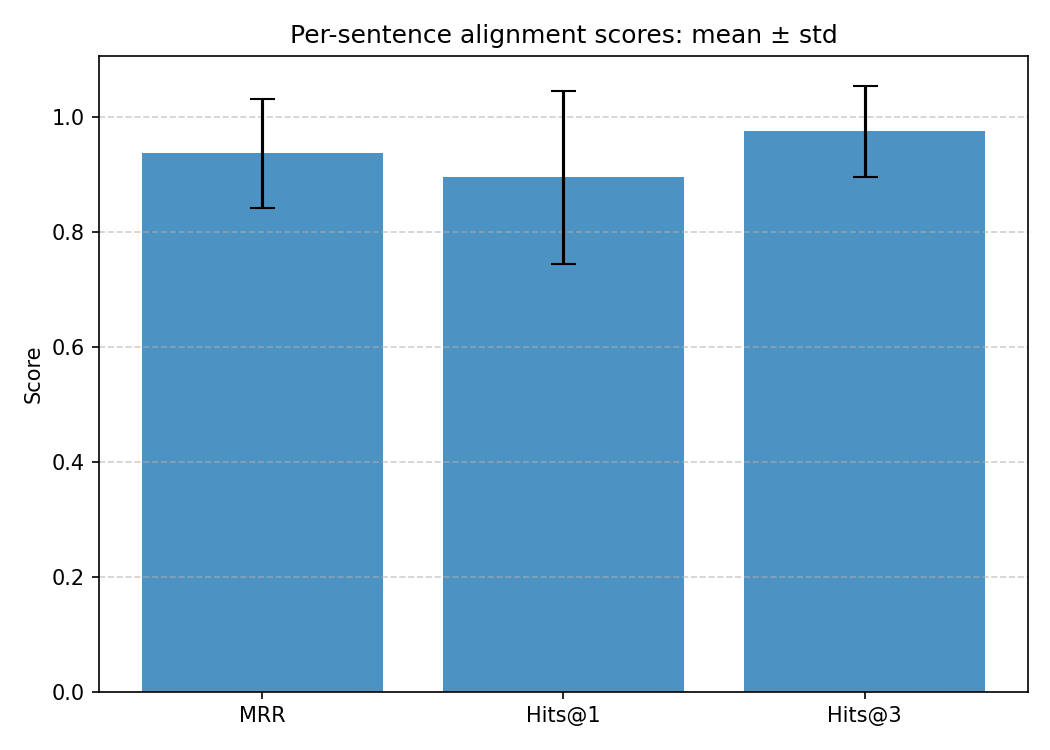}
    \caption{\footnotesize{\textbf{OT and translation.} 
Average MRR (defined in~\eqref{eq:mrr}), Hits@1, and Hits@3 (defined in~\eqref{eq:hits}) computed over $10{,}000$ sentences from the WMT14 English–French dataset using OT between word embeddings. Error bars mark the standard deviation across sentences. The OT solution aligns almost $90\%$ of the translated word pairs. }}
    \label{fig:OT_embeddings}
\end{figure}

We compute OT solution with entropy regularization parameter $\lambda = 0.05$~\citep{cuturi2013sinkhorn} using the POT library~\citep{flamary2021pot}. 
Fig.~\ref{fig:OT_embeddings} shows that OT aligns translated words with high accuracy: the average Hits@1 across the $10{,}000$ sentence pairs is nearly $0.9$, indicating that OT correctly identifies about $90\%$ of translation pairs.

\section{Discussions}

We studied the mechanism of token alignment, with optimal transport (OT), in attention layers. We experimentally link token alignment to language translation, showing that attention weights can iteratively match semantically equivalent words, iteratively across the layers. We then analyzed the underlying mechanism of OT in transformers, proving that attention layers can approximate the OT alignment up to an error that vanishes with depth.

Our findings open several promising avenues for future research. In particular, we outline four directions that can deepen our understanding of token alignment mechanisms in large language models.

\emph{Depth Efficient Guarantees.} According to Theorem~\ref{thm:convergence}, a transformer with depth \(O(\epsilon^{-2})\) can achieve an \(O(\epsilon)\)-accurate solution, following the established convergence rate for gradient descent with adaptive step sizes. However, only \(O(\log(1/\epsilon))\) Sinkhorn iterations are needed for the same accuracy. We believe this gap arises from a loose convergence analysis, which can be refined in future work.

\emph{Learning with Small Prompts.} We prove that a transformer with fixed parameters can solve OT for any arbitrary $n$. This striking generalization has practical benefits: it can drastically reduce both training time and memory usage for assignment tasks, since the transformer can be trained on prompts with a constant number of tokens. A natural question arises: what is the minimal value of $n$ sufficient for the model to learn the assignment task?

\emph{Prompt Engineering.} We theoretically and experimentally demonstrate that prompt engineering is essential for in-context assignment. However, the underlying mechanism of prompt engineering remains understudied in a broader context. Our findings motivate further study of prompt engineering from a computational perspective, highlighting its role in enhancing the computational expressivity of transformers.

\emph{Generalized Cost Functions.}
The OT problem can be formulated with a general cost function, beyond the standard Euclidean distance \( \|x_i - y_j\|^2 \). An important question is whether standard softmax attention can solve the assignment problem with a general cost function \( C(x_i, y_j) \). We conjecture that a combination of attention layers and feedforward networks can approximate solutions in general cases. 
\bibliography{references} 
\clearpage
\appendix
\thispagestyle{empty}

\onecolumn
\newpage
\newpage
\onecolumn

\begin{center}
   \LARGE{Appendix}
\end{center}
\section{Experimental settings} \label{sec:experiments_app}

\paragraph{Training loss.}
 We generate random data and train a transformer to solve the assignment task by minimizing
\begin{equation}\begin{aligned}\label{eq:training_objective}
\includegraphics[width=0.5\linewidth]{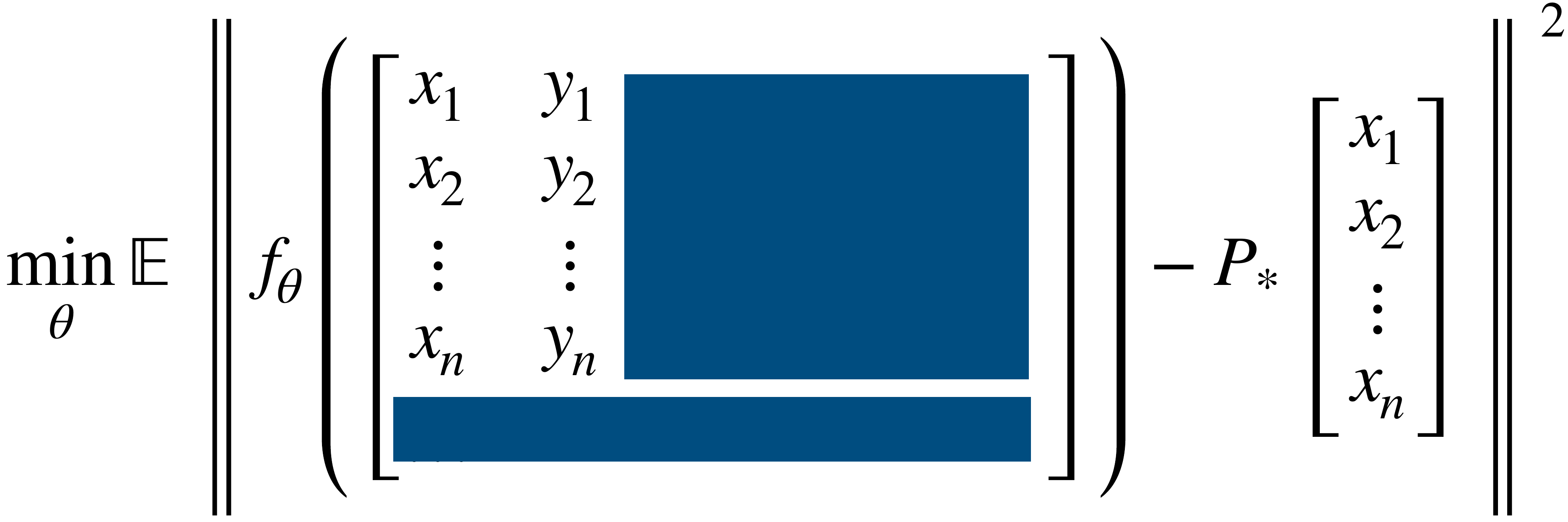}
\end{aligned},
\end{equation} 
where the expectation is taken over randomly generated inputs $x_1, \dots, x_n$, and the \textcolor{blue}{$\blacksquare$} marks engineered part of the input in \eqref{eq:engineered_prompt}. $f_\theta:\R^{(n+1)\times (2d+9)} \to \R^{n}$ is the output function of a transformer with parameters $\theta$. To generate outputs, we use an attention layer defined as 
\begin{align}
   f_\theta(Z_0) =  [\attn_{\theta}([Z_{\ell}]_{1:n,:})]_{:,0:d}
\end{align}
where $[Z_\ell]_{1:n,:}$ denotes the first $n$ rows of $Z_{\ell}$ in \eqref{eq:Zdynamics}. 
The above indexing allows us to generate the outputs of size $n\times d$. 

\paragraph{Training data.} The expectation in the training loss~\eqref{eq:training_objective} is taken over 1000 random samples generated with $n=7$. $x_1, \dots, x_n$ are a random permutation of $[\sfrac{1}{n}, \sfrac{2}{n}, \dots, \sfrac{n}{n}]$, and we set $y_i = \sfrac{i}{n}$ in our experiments.

\paragraph{Optimization.} For training, we run $10^4$ iterations of Adam~\citep{kingma2014adam} with stepsize 0.001. Parameters are initialized from a Gaussian distribution with covariance matrix $\sfrac{1}{(2d+9)}$. We set $B_j^{(\ell)} = (\sfrac{1}{20})I_{2d+9}$ where $I_k$ is the square identity matrix of size $k$. We optimize the attention parameters $\theta := \{ [K^{(m,j)}, Q^{(m,j)}, V^{(m,j)}]\}_{m=1}^\ell$ using the reparameterization $ P^{(\ell,j)}=K^{(\ell,j)} Q^{(\ell,j)}$ and optimize $P^{(\ell,j)}$.

\paragraph{Without prompt engineering.} \label{section:prompt}
To study the impact of prompt engineering in Fig.~\ref{fig:train} Observation~(2), we remove additional columns in the engineered prompt as
\begin{align} \label{eq:woengineering}
    Z' = \begin{bmatrix}
    x_1 & y_1 &  0\\
    \vdots & \vdots & \vdots  \\
    x_n & y_n &  0 \end{bmatrix}   \in \R^{n \times 3}.
\end{align}   
We trained the model on exactly the same training data with the same optimization settings. 
\paragraph{Computational resources.}
All experiments are implemented in PyTorch~\citep{paszke2019pytorch} and executed on a NIVIDA RTX 6000 Ada GPU. We also used POT library~\citep{flamary2021pot} to compute $P^*$ and $P^*_\lambda$.
Our implementation is included in the supplementary material. 
\paragraph{GenAI Statement.}
Figures~\ref{fig:translation},~\ref{fig:attention_algiment},~\ref{fig:attention_algiment_de}, and~\ref{fig:OT_embeddings} present experiments implemented with the assistance of GPT-5, while the illustration in Figure~\ref{fig:sketch} was created using GPT-4o.
\section{The Aligning Transformer} \label{sec:aligning_tr}
The theoretical and experimental results presented are based on a specific choice of parameters for the attention layers. These parameters are used both to generate the plots in figures~\ref{fig:convergence} and \ref{fig:n} and also used to prove the main theorems.

Recall that the hidden representations in the transformer obeys 
\begin{equation*} Z_{\ell+1} = Z_{\ell}+ \sum_{j=1}^2 \attn_{\theta_j^{(\ell)}}(Z_{\ell}) B_j^{(\ell)}, 
\end{equation*}
where \(\{\theta_j^{(\ell)}\} \) denotes the parameters of attention head $j$ at layer $\ell$ containing three matrices as \( \theta_j^{(\ell)} = \{K^{(\ell,j)}, Q^{(\ell,j)}, V^{(\ell,j)} \in \R^{d\times d}\} \). 

We define \( P^{(\ell,j)} = K^{(\ell,j)} (Q^{(\ell,j)})^\top \). Let \( p = d + 9 \), and let \( e_i^{(k)} \in \mathbb{R}^{k} \) denote the \( i \)-th standard basis vector, defined by
$
[e_i]_j = \begin{cases}
1 & \text{if } i = j, \\
0 & \text{otherwise}
\end{cases}.
$
Based on these definitions, Table~\ref{tab:parameters} summarizes the choice of parameters for OT.  
\begin{table}[]
    \centering
    \begin{tabular}{c|c}
    \hline
    \hline \\ 
    & \\
        $\lambda P^{(\ell,1)}$ & $\begin{bmatrix}
\begin{bmatrix}
    0_{d\times d} \\
    0_{p\times d}
\end{bmatrix}   & 2\begin{bmatrix}
    I_{d\times d} \\
    0_{p\times d}
\end{bmatrix}   & \begin{bmatrix}
    0_d \\
    0_{p}
\end{bmatrix} &  \begin{bmatrix}
0_d \\
 -e_{d+3}^{(p)}
\end{bmatrix}& \begin{bmatrix}
0_d \\
-e_{d+1}^{(p)}
\end{bmatrix} & \begin{bmatrix}
0_d \\
 e_{d+7}^{(p)}
\end{bmatrix} & \begin{bmatrix}
0_d \\
 -\lambda e_{d+5}^{(p)}
\end{bmatrix}  & \begin{bmatrix}
    0_{d\times 2} \\
    0_{p\times 2}^{(p)}
\end{bmatrix}   & \begin{bmatrix}
0_d \\
 e_{2d+5}^{(p)}
\end{bmatrix} & \begin{bmatrix}
    0_{d} \\
    0_{p}
\end{bmatrix}
    \end{bmatrix}$  \\
    &  \\
    \hline  \\
   $\lambda P^{(\ell,2)}$     &  $ \begin{bmatrix}
    2\begin{bmatrix}
        0_{d\times d} \\
        I_{d\times d} \\
        0_{9\times d}
    \end{bmatrix}
         &  \begin{bmatrix}
        0_{d\times d} \\
        0_{d\times d} \\
        0_{9\times d}
    \end{bmatrix}& \begin{bmatrix}
        0_{d} \\
        I_{d} \\
        -e_{3}^{(9)}
    \end{bmatrix} & \begin{bmatrix}
        0_{d} \\
        0_{d} \\
        0_3
    \end{bmatrix} & \begin{bmatrix}
        0_{d} \\
        I_{d} \\
        -e_{2}^{(9)}
    \end{bmatrix}& \begin{bmatrix}
        0_{d} \\
        I_{d} \\
        e_{8}^{(9)}
    \end{bmatrix} &  \begin{bmatrix}
        0_{d} \\
        I_{d} \\
        -\lambda e_{5}^{(9)}
    \end{bmatrix} & \begin{bmatrix}
        0_{d} \\
        0_{d} \\
        0_3
    \end{bmatrix} & \begin{bmatrix}
        0_{d} \\
        I_{d} \\
         e_{5}^{(9)}
    \end{bmatrix} & \begin{bmatrix}
        0_{d\times 2} \\
        0_{d\times 2} \\
        0_{3\times 2}
    \end{bmatrix}
    \end{bmatrix}$ \\
    & \\ 
    \hline \\ 
    $[V^{(\ell,1)}]_{ij}$ & $=\begin{cases}
        1 & i = 2d + 6 
 \; \text{and} \; j=2d + 7 \\ 
        0 & \text{otherwise}
    \end{cases}$ \\
    & \\
    \hline \\ 
    $[V^{(\ell,2)}]_{ij}$ & $= \begin{cases}
        1 & i = 2d+6 
 \; \text{and} \; j=2d+8 \\ 
        0 & \text{otherwise}
    \end{cases}$ \\
    & \\
    \hline\\
    $ B_j^{(\ell)}$ & $\gamma_\ell I_{(2d+9)\times (2d+9)}$ \\
    & \\
    \hline
    \hline
    \end{tabular}
    \caption{\textbf{Parameters for OT:} Our proof is constructive since it specifies the choice of parameters.}
    \label{tab:parameters}
\end{table}

\section{Proof of Theorem~\ref{thm:gd}}
We demonstrate that two attention heads can jointly implement a single step of gradient descent (with adaptive step sizes) on \(L(u, v)\). By induction, multiple attention heads can implement several iterations of gradient descent with adaptive step sizes. The proof is constructive, explicitly determining the choice of parameters specified in Section~\ref{sec:aligning_tr}.

Recall $u_\ell,v_\ell \in \R^n$ are iterations of gradient descent (with adaptive coordinate-wise stepsize) on Lagrangian function $L$ defined in~\eqref{eq:gd}:
\begin{align*}
\begin{cases}
    u_{\ell+1} = u_{\ell} -  D_{\ell} \nabla_u L(u_{\ell},v_{\ell}) \\ 
    v_{\ell+1} = v_{\ell} - D_{\ell}' \nabla_v L(u_{\ell},v_{\ell})
\end{cases},  L(u,v) := \lambda \left( \sum_{ij} e^{(-C_{ij} + u_i +v_j)/\lambda-1}  \right)  - \sum v_i - \sum_{i} u_i 
\end{align*}
where the coordinate-wise stepsizes are
\begin{align*}
    (\gamma_{\ell}[D_{\ell}]_{ii})^{-1} =  \sum_{j} e^{(-C_{ij}+[u_\ell]_i+[v_\ell]_{j})/\lambda-1}+1, \quad  (\gamma_{\ell}[D_{\ell}']_{jj})^{-1} =  \sum_{i} e^{(-C_{ij}+[u_\ell]_i +[v_\ell]_j)/\lambda-1}+1. 
    \end{align*}

Define $x = [x_1, \dots, x_n], y = [y_1,\dots, y_n] \in \R^{n\times d}$ and $x^2 = [\|x_1\|^2, \dots, \|x_n\|^2], y^2 = [\|y_1\|^2, \dots, \| y_n \|^2] \in \R^n$. Let $H^{(\ell)}\in \R^{n\times n}$ obeys 
\begin{align}
    H^{(\ell)} = \tfrac{1}{\lambda} \left(-x^2  \ones_n^\top + 2 x y^\top - \ones_n  (y^2)^\top + u_{\ell}  \ones_n^\top + \ones_n  (v_{\ell})^\top \right) - \ones_n  \ones_n^\top 
\end{align}
Define $M^{(\ell)} \in \R^{n\times n}$ defines as $M_{ij}^{(\ell)} = e^{H_{ij}^{(\ell)}}$. Gradient descent thus follows:  
\begin{align}
    u_{\ell+1} &= u_\ell - D_{\ell} (M^{(\ell)} \ones_n -\ones_n) \\
    v_{\ell+1} &= v_\ell - D'_{\ell} ((M^{(\ell)})^\top \ones_n - \ones_n) \label{eq:gd_iterates_closed_form}
\end{align}

\paragraph{Induction statement.} We assume the statement holds for $\ell$ and then prove it for $\ell+1$. Thus, the induction hypothesis is
\begin{align*}
    Z_\ell = \begin{bmatrix}
    \textcolor{blue}{x_1} &  \textcolor{blue}{y_1} & \textcolor{red}{\|x_1\|^2} & \textcolor{red}{\|y_1\|^2} & \textcolor{red}{1_4} & \textcolor{teal}{[u_{\ell}]_1} & \textcolor{teal}{[v_{\ell}]_1} & \textcolor{red}{0} \\ 
    \textcolor{blue}{x_2} &  \textcolor{blue}{y_2} & \textcolor{red}{\|x_2\|^2} & \textcolor{red}{\|y_2\|^2} & \textcolor{red}{1_4} & \textcolor{teal}{[u_{\ell}]_2} & \textcolor{teal}{[v_{\ell}]_2} & \textcolor{red}{0}\\
    \textcolor{blue}{\vdots} & \textcolor{blue}{\vdots} & \textcolor{red}{\vdots} & \textcolor{red}{\vdots}& \textcolor{red}{\vdots}& \textcolor{red}{\vdots} & \textcolor{teal}{\vdots} & \textcolor{red}{\vdots}  \\
    \textcolor{blue}{x_n} &  \textcolor{blue}{y_n} & \textcolor{red}{\|x_n\|^2} & \textcolor{red}{\|y_n\|^2} & \textcolor{red}{1_4}& \textcolor{teal}{[u_\ell]_d} & \textcolor{teal}{[v_\ell]_d} & \textcolor{red}{0} \\
      \textcolor{blue}{0} & \textcolor{blue}{0} & \textcolor{red}{0} & \textcolor{red}{0} & \textcolor{red}{-v_4} &  \textcolor{teal}{?}  & \textcolor{teal}{?} & \textcolor{red}{0}  
    \end{bmatrix}   \in \R^{(n+1) \times (2d+9)},
\end{align*}
where $1_4$ denotes the all-ones 4-dimensional vector, $v_4 = [0, 0, 0, 1]$, and $[u]_{i}$ denotes element $i$ of vector $u$. It is easy to check that the above equation holds for $\ell=0$ as $u_0= v_0 = 0_{n}$. 
 The choice of $w_v^{(\ell,j)}$ ensure that only the $2d+7$-th and $2d+8$-th columns of $Z_{\ell}$ change with $\ell$, which are highlighted in \textcolor{teal}.
 We prove that
\begin{align*}
     Z_{\ell+1} = \begin{bmatrix}
    \textcolor{blue}{x_1} &  \textcolor{blue}{y_1} & \textcolor{red}{\|x_1\|^2} & \textcolor{red}{\|y_1\|^2} & \textcolor{red}{1_4} & \textcolor{teal}{[u_{\ell+1}]_1} & \textcolor{teal}{[v_{\ell+1}]_1} & \textcolor{red}{0} \\ 
    \textcolor{blue}{x_2} &  \textcolor{blue}{y_2} & \textcolor{red}{\|x_2\|^2} & \textcolor{red}{\|y_2\|^2} & \textcolor{red}{1_4} & \textcolor{teal}{[u_{\ell+1}]_2} & \textcolor{teal}{[v_{\ell+1}]_2} & \textcolor{red}{0}\\
    \textcolor{blue}{\vdots} & \textcolor{blue}{\vdots} & \textcolor{red}{\vdots} & \textcolor{red}{\vdots}& \textcolor{red}{\vdots}& \textcolor{red}{\vdots} & \textcolor{teal}{\vdots} & \textcolor{red}{\vdots}  \\
    \textcolor{blue}{x_n} &  \textcolor{blue}{y_n} & \textcolor{red}{\|x_n\|^2} & \textcolor{red}{\|y_n\|^2} & \textcolor{red}{1_4}& \textcolor{teal}{[u_{\ell+1}]_d} & \textcolor{teal}{[v_{\ell+1}]_d} & \textcolor{red}{0} \\
      \textcolor{blue}{0} & \textcolor{blue}{0} & \textcolor{red}{0} & \textcolor{red}{0} & \textcolor{red}{-v_4} &  \textcolor{teal}{?}  & \textcolor{teal}{?} & \textcolor{red}{0}  
    \end{bmatrix}   \in \R^{(n+1) \times (2d+9)},
\end{align*}

Indeed, the extended prompt offers sufficient memory to store the iterates of gradient descent.

\paragraph{Induction proof.}  We begin by computing the output of the first attention head in layer $\ell+1$, step by step. Through straightforward algebra, we obtain the following: 
    \begin{align*}
    Z_{\ell} P^{(\ell,1)} = \begin{bmatrix}
        0 & 2x/\lambda & 0_n& -\ones_n/\lambda  & -\|x\|^2/\lambda & u_{\ell}/\lambda &  -1_n & 0_{2\times n} & 1_n/\lambda& 0  \\ 
        0 & 0 & 0 & 0 & 0 & 0 & 0 & \zeros_{2\times 1} & 0 & 0 
    \end{bmatrix} \in \R^{(n+1) \times d'}
\end{align*}
where matrices $x$ and $y$ are defined earlier in the proof. With these notations and the preceding equation, we proceed as follows:
\begin{align*}
     Z_\ell P^{(\ell,1)} Z_\ell^\top  
    & =  \begin{bmatrix}
         H^{(\ell)} & 0_{n} \\ 
         0_{n}^\top & 0
    \end{bmatrix}
\end{align*}
which obtains 
\begin{align}
    \exp(Z_{\ell} P^{(\ell,1)} Z_\ell^\top) = \begin{bmatrix}
        M^{(\ell)} & \ones_n \\ 
        \ones_n^\top & 1
    \end{bmatrix}
\end{align}

Furthermore, the choice of parameters $V^{(\ell,1)}$ obtains 
\begin{align*}
    Z_\ell V^{(\ell,1)} = -\gamma \begin{bmatrix}
        \zeros_{n} & \dots &  \zeros_n & \ones_n & \zeros_n & \zeros_n \\ 
        0 & \dots &  0 & -1 & 0 & 0
    \end{bmatrix}
\end{align*}
Stitching all equations together yields  
\begin{align*}
\attn_{\theta^{(\ell)}_1} (Z_\ell) B_1^{(\ell)} = \begin{bmatrix}
        \zeros_{n} & \dots &  \zeros_n & - D_{\ell} (M^{(\ell)} \ones_n - \ones_n) & \zeros_n & \zeros_n \\ 
        0 & \vdots &  0 & n-1 & 0 & 0
    \end{bmatrix}
\end{align*}  Observe that the matrix above contains the gradient $\nabla_{u} L(u_\ell, v_\ell)$ from Eq.~\eqref{eq:gd_iterates_closed_form}, which is required to compute the next gradient descent iterate $u_{\ell+1}$. 
Similarly, we can show that 
\begin{align*}
\attn_{\theta^{(\ell)}_2} (Z_{\ell}) B_2^{(\ell)} = \begin{bmatrix}
        \zeros_{n} & \dots &  \zeros_n  & \zeros_n & - D'_{\ell} (M^{(\ell)})^\top \ones_n - \ones_n)& \zeros_n \\ 
        0 & \vdots &  0 & 0 & n-1 & 0
    \end{bmatrix},
\end{align*}
which computes $v_{\ell+1}$ in parallel using a second attention head. Thus, 
substituting the last two equations into \eqref{eq:Zdynamics} concludes the induction proof.

\section{ Theorem~\ref{thm:convergence}} \label{app:convergence}
\subsection{Formal statement}
Attention matrices can approximate the entropy-regularized assignment solution, denoted by \( P^*_\lambda \), as defined in~\eqref{eq:ot_reg}. Consider the block of \emph{attention matrices} \( A^{(\ell)} \in \mathbb{R}^{n \times n} \), defined by
\begin{align} \label{eq:A}
   A_{ij}^{(\ell)} = \frac{e^{\langle K^{(\ell,1)} z_i^{(\ell)}, Q^{(\ell,1)} z_j^{(\ell)} \rangle}}{\sum_{j=1}^n e^{\langle K^{(\ell,1)} z_i^{(\ell)}, Q^{(\ell,1)} z_j^{(\ell)} \rangle} }
\end{align}
where \( z_i^{(\ell)} \) is the \( i \)-th row of \( Z_{\ell} \), representing the token embeddings at layer \( \ell \), and \( Q^{(\ell,1)} \), \( K^{(\ell,1)} \) are the query and key weight matrices of an attention head at layer \( \ell \).

We prove the attention matrix \( A^{(\ell)} \) converges to \( P^*_\lambda \) in an appropriate metric for certain choice of parameters of attention layers. As discussed, solution $P^*_\lambda$ in~\eqref{eq:ot_reg} can be written as 
\begin{equation}
\begin{aligned} \label{eq:Q}
    P^*_\lambda &= \diag(p^*) Q \diag(q^*) \\ 
    & \quad p^*,q^* \in \R^n_+, \quad  Q \in \R^{n\times n}_+,
\end{aligned}
\end{equation}
where $\diag(v)$ is a diagonal matrix whose diagonal element in row $i$ is $v_i$, $\quad Q_{ij} = e^{-\tfrac{C_{ij}}{\lambda} -1}$, $p^*_i = e^{v_i^*/\lambda}$ and $q^*_j = e^{u_j^*/\lambda}$. 
It is easy to check that replacing $p^*$ and $q^*$ with $c p^*$ and $q^*/c$ leads to the same matrix $P^*_{\lambda}$ for all $c \in \R_+$. \citet{franklin1989scaling} introduce a metric that accounts for this particular scaling invariance:
\begin{align} \label{eq:d}
\mu(q,q') = \log\left( \max_{ij} \frac{q_iq'_j}{q_j q'_i}\right).
\end{align}
Remarkably,  $\mu$ is a metric that satisfies the triangle inequality \citep{franklin1989scaling}. However, $\mu$ is not a norm, as $\mu(q,q')=0$ only implies that there exists a constant $c$ such that $q=c q'$. The next theorem establishes an explicit convergence rate for the attention matrices $A^{(\ell)}$ to $P_\lambda^*$ in $\mu$.

\begin{theorem}[Formal version of Thm.~3.2: Convergence to $P_\lambda^*$ in the Hilbert projective metric]
\label{thm:formal}
Let $C \in \mathbb{R}^{n \times n}$ be a cost matrix with entries $C_{ij}$, and fix $\lambda>0$.
Define
\[
Q \;\in\; \mathbb{R}^{n\times n}_{+},
\qquad
Q_{ij} \;=\; e^{-\,C_{ij}/\lambda-1}.
\]
Let $P_\lambda^* \in \mathbb{R}^{n\times n}_{+}$ denote the entropy-regularized optimal transport plan,
which admits the factorization
\[
P_\lambda^* \;=\; \mathrm{diag}(p^*)\, Q \,\mathrm{diag}(q^*)
\quad \text{for some } p^*,q^* \in \mathbb{R}^n_{+},
\]
unique up to positive rescaling of $p^*,q^*$. Consider a transformer with a fixed choice of
parameters (independent of $n$, $d$, and the input) as constructed in Appendix~B, and let
$A^{(\ell)} \in \mathbb{R}^{n\times n}_{+}$ denote the attention matrix (for a specified head/block) at depth $\ell$.
For each $\ell \ge 1$ there exist $p_\ell,q_\ell \in \mathbb{R}^n_{+}$ such that
\[
A^{(\ell)} \;=\; \mathrm{diag}(p_\ell)\, Q \,\mathrm{diag}(q_\ell).
\]

Define the Hilbert projective metric
\[
\mu(u,v) \;=\; \log \!\left( \max_{i,j} \frac{u_i v_j}{u_j v_i} \right),
\qquad u,v \in \mathbb{R}^n_{+},
\]
and let
\[
\phi(Q) \;=\; \max_{i,j,k,\ell} \frac{Q_{ik}\,Q_{j\ell}}{Q_{jk}\,Q_{i\ell}},
\qquad
\eta \;=\; \frac{\sqrt{\phi(Q)}-1}{\sqrt{\phi(Q)}+1} \in [0,1).
\]
Set
\[
\frac{1}{4}\, r^2 \;=\; \|p_1 - p^*\|_2^2 \;+\; \|q_1 - q^*\|_2^2 .
\]
Then, provided $\ell \;\ge\; 64\,n\,e^{3r/\lambda}\,r$, there exists an index $k \le \ell$ such that
\[
\min_{k \le \ell}\;
\max\bigl\{\,\mu(p_k,p^*),\;\mu(q_k,q^*)\,\bigr\}
\;\le\;
\frac{36 \sqrt{n} \, e^{1.5 r/\lambda}}{\sqrt{\ell}\,(1-\eta)} \,.
\]
\end{theorem}
\begin{remark}
We improved the bound in Thm.~\ref{thm:convergence} from $O\bigl(\tfrac{n^{3/2}}{\ell^{1/2}}\bigr) \to O\bigl(\sqrt{\tfrac{n}{\ell}}\bigr)$ while preparing the supplementary material.

\end{remark}

\begin{remark}[On scaling, metric choice, and non-monotonicity]
\label{rem:scaling_metric}
(i) The factorization of $P_\lambda^*$ is invariant to the rescaling $(p^*,q^*) \mapsto (c\,p^*, c^{-1} q^*)$, $c>0$;
the metric $\mu$ is designed to quotient out this invariance. 
(ii) The ``$\min_{k\le \ell}$'' accounts for possible non-monotone progress across layers; the bound holds for the
best iterate up to depth $\ell$. 
(iii) The constant $r$ encodes the initialization distance to $(p^*,q^*)$, and the quantity $\eta$ is the classical
Sinkhorn contraction factor \citep{franklin1989scaling}.
\end{remark}
\subsection{Proof}
\paragraph{Challenge:}According to Thm.~\ref{thm:gd}, a transformer can implement gradient descent. Therefore, the proof casts to analyzing gradient descent (with specific coordinate-wise stepsizes) on the convex $L$. However, we cannot directly apply existing convergence results from convex optimization. The existing convergence results for smooth convex optimization are in terms of function value $L$ when $L$ is not strongly convex\footnote{It is easy to check that $L$ is not strongly convex since it Hessian has a zero eigenvalue.}. But, the theorem statement aims at the convergence to a minimizer. 

\begin{figure}[h]
    \centering
    \includegraphics[width=0.5\linewidth]{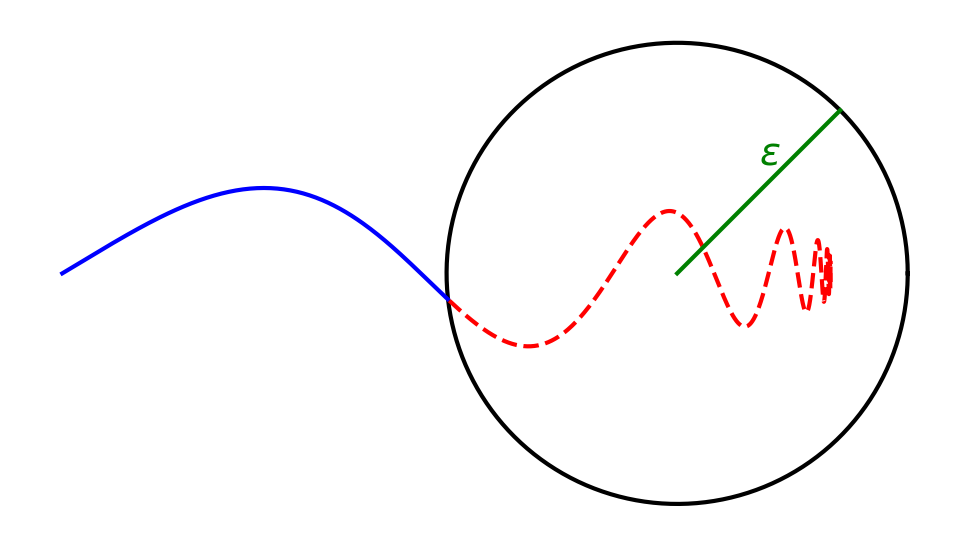}
    \caption{\footnotesize{\textbf{Proof techinque for Theorem~\ref{thm:convergence}.} We first prove that the attention matrix converges to a local neighborhood of the set of doubly stochastic matrices. This convergence is illustrated by the blue curves converging to a small circle. Next, we show that this convergence implies convergence to the minimizer. To establish this, we hypothetically run Sinkhorn's iterations and leverage their known convergence rate. The red curve illustrates these hypothetical Sinkhorn iterations. 
  }}
    \label{fig:sketch}
\end{figure}

\paragraph{Proof Sketch.}
 The proof consists of two key steps: (i) the convergence of attention matrix $A^{(\ell)}$ to a matrix that is approximately doubly stochastic, and (ii) a hypothetical simulation of Sinkhorn's recurrence. See Figure~\ref{fig:sketch} for the illustration.
\begin{itemize}
    \item[(i)] As shown in Thm.~\ref{thm:gd}, the transformer can perform gradient descent with an adaptive step size on the convex function $L$. Since $L$ is convex, gradient descent is guaranteed to converge to a stationary point where the gradient norm becomes zero. Specifically, it is straightforward to verify that $\nabla_u L = M \mathbf{1} -  \mathbf{1}$ and $\nabla_v L = M^\top \mathbf{1} -  \mathbf{1}$, where $M_{ij} = \exp\left( \tfrac{-C_{ij} + u_i + v_j}{\lambda} - 1 \right)$. Therefore, small gradients for $u$ and $v$ imply that $M$ is close to being doubly stochastic.  
    \item[(ii)] We demonstrate that when the matrix $M$ is approximately doubly stochastic, it is near the desired solution $P^*_\lambda$. To establish this, we (hypothetically) run Sinkhorn’s recurrence starting from $M$ and use its contraction property proven by \citep{franklin1989scaling}.
\end{itemize}
Before elaborating on the details of (i) and (ii), we present two propositions.
\paragraph{Preliminaries.}
Define the functions $row: \R^{n \times n}_+ \to \R^n_+$ and $col: \R^{n \times n}_+ \to \R^n_+$ as
\begin{align*}
    \row(A)_i = \frac{1}{ \sum_{j} A_{ij}}, \quad \col(A)_j = \frac{1}{ \sum_{i} A_{ij}}.
\end{align*}
We also introduce functions $f, g: \R^{n \times n} \to \R^{n \times n}$ defined as
\begin{align*}
    f(A) = A \diag(\col(A)), \quad g(A) = \diag(\row(A)) A.
\end{align*}
Indeed, $f(A)$ (resp. $g(A)$) normalizes the columns (resp. rows) of $A$ by a scaling factor of their average. We will later use $f$ and $g$ to formulate Sinkhorn’s recurrence, which iteratively normalizes the rows and columns of a matrix with positive elements. The next proposition proves that an almost doubly stochastic matrix remains almost doubly stochastic under $f$ and $g$. To formulate the statement, we introduce a set containing matrices that almost doubly stochastic matrices:
\[
\S_\epsilon := \left\{ A \in \R^{n\times n}_+ \mid \| A \ones_n - \ones_n \|_{\infty} \leq \epsilon \text{ and } \| A^\top \ones_n - \ones_n \|_{\infty} \leq \epsilon \right\}.
\]

\begin{proposition} \label{prop:S}
    Suppose that $A \in \S_\epsilon$; then $f(A) \in \S_{3\epsilon}$ and $g(A) \in \S_{3\epsilon}$, as long as $\epsilon < \sfrac{1}{3}$.
\end{proposition}

Recall the metric $d$ defined in \eqref{eq:d}. The next proposition establishes a particular property of $f$ and $g$ with respect to $d$.
 
\begin{proposition} \label{prop:d_bound}
   Let $A \in \S_{\epsilon}$ be decomposed as $A = \diag(w) Q \diag(q)$, where $w, q \in \mathbb{R}_+^n$.

\begin{itemize}
    \item[(i)] For $f(A) = \diag(w) Q \diag(q')$,  $\mu(q, q') \leq 4 \epsilon$ holds for $\epsilon < \tfrac{1}{4}$.
    \item[(ii)] For $g(A) = \diag(w') Q \diag(q)$,  $\mu(w, w') \leq 4 \epsilon$ holds for $\epsilon < \tfrac{1}{4}$.
\end{itemize}
\end{proposition}

\paragraph{Convergence Analysis.}
According to Theorem~\ref{thm:gd}, there is a choice of parameters such that  
\[
A_{ij}^{(\ell)} = e^{ \frac{-C_{ij} + [u_{\ell}]_i + [v_{\ell}]_j}{\lambda} - 1 },
\]
where \( u_{\ell} \) and \( v_{\ell} \) are the iterates defined in \eqref{eq:gd}. The following lemma establishes that, as the number of iterations \( \ell \) increases, $A^{(\ell)}$ meets a neighborhood of doubly stochastic matrices.

\begin{lemma} \label{lemma:convergence}
For $\gamma_k^{-1} =(n+2) e^{\sfrac{2r}{\lambda}}$, there exists a \( k \leq \ell \) such that
    \begin{align*}
     A^{(k)} \in \S_{\epsilon}, \quad \text{ where } \epsilon^2 := \left(\frac{1}{\ell}\right) 3n e^{\sfrac{3r}{\lambda}}r.
    \end{align*}
\end{lemma}

Notably, the matrix \( A^{(k)} \) has a specific structure that ensures \( A^{(k)} \in \S_\epsilon\) is sufficient to approximate \( P^*_\lambda \). To prove this statement, we leverage the contraction property of Sinkhorn's recurrence.

\paragraph{Contractive Sinkhorn's Process.}  
According to the last lemma, there exists an iteration \( k \leq \ell \) such that \( A^{(k)} \in \mathcal{S}_{\epsilon} \). We then apply Sinkhorn's recurrence starting from \( A_1 = g(A^{(k)}) \) as:
\[
A_{m+1/2} = f(A_m), \quad A_{m+1} = g(A_{m+1/2}).
\]
 Notably, we utilize the above recurrence solely for the proof; hence, there is no need for a transformer to implement this recurrence. According to the definition, \( A_m \) can be decomposed as \( \diag(w_m) Q \diag(q_m) \), where \( Q_{ij} = e^{-C_{ij}/\lambda - 1} \) and \( q_m, w_m \in \mathbb{R}^n_+ \). \cite{sinkhorn1967diagonal} proves that there exist unique vectors \( w^*, q^* \in \mathbb{R}^n_+ \) such that \( P^*_\lambda = \diag(w^*) Q \diag(q^*) \), where \( w^*_i = e^{u^*_i/\lambda} \) and \( q^*_j = e^{v^*_j/\lambda} \). \citep{franklin1989scaling} establish the linear convergence of \( (w_m,q_m) \) to \( (w^*, q^*) \):
\begin{align} \label{eq:contract}
   \begin{cases}
      \mu(w_{m+1}, w^*) \leq \eta \mu(w_m, w^*)  \\ 
      \mu(q_{m+1}, q^*) \leq \eta \mu(q_m, q^*) 
   \end{cases}, \quad \eta  = \frac{\phi(A_1)^{1/2}-1}{\phi(A_1)^{1/2}+1} < 1, 
\end{align}
where \( \phi(A) = \max_{ijkl}\frac{A_{ik}A_{jl}}{A_{jk} A_{il}} \). Since \( A_1 = \diag(w_1) Q \diag(q_1) \), we have \( \phi(A_1) = \phi(Q) \).

Propositions~\ref{prop:S} and \ref{prop:d_bound} enable us to demonstrate that there exists a constant \( c \) such that \( cA_1 \) lies within a neighborhood of \( P^*_\lambda \). 
Proposition~\ref{prop:S} combined with Lemma~\ref{lemma:convergence} ensure $A_1  \in \mathcal{S}_{3\epsilon}$. Thus, we can apply Proposition~\ref{prop:d_bound} to obtain:
$ \mu(q_{2}, q_{1}) \leq 12 \epsilon.
$
Using Proposition~\ref{prop:S} again, we find that \( A_{1 + 1/2} \in \mathcal{S}_{9\epsilon} \). Consequently, we can invoke Proposition~\ref{prop:d_bound} once more to yield:
$
    \mu(w_{2}, w_{1}) \leq 36 \epsilon.
$ Applying the triangle inequality together with \ref{eq:contract} completes the proof:
\begin{align*}
    36 \epsilon &\geq \mu(w_{2}, w_{1}) \geq \mu(w_1, w^*) - \mu(w_{2}, w^*) \geq (1-\eta) \mu(w_{1}, w^*) \\ 
    12  \epsilon &\geq \mu(q_{2}, q_{1}) \geq \mu(q_1, q^*) - \mu(q_{2}, q^*) \geq (1-\eta) \mu(q_{1}, q^*).
\end{align*}
\section{Proof of Lemma~\ref{lemma:convergence}}

\paragraph{Notations.}
Define the concatenated vector of iterates as
\[
\theta_{k} = \begin{bmatrix} u_{k} \\ v_{k} \end{bmatrix},
\]
and consider the following block diagonal matrix:
\[
\Lambda_k = \begin{bmatrix}
    D_k & 0 \\ 
    0 & D'_k
\end{bmatrix},
\]
where \( D_k \) and \( D'_k \) are diagonal matrices at iteration \( k \) defined in Theorem~\ref{thm:gd}.
Define the ball \( B(r) = \{ \theta \in \mathbb{R}^n \mid \| \theta \| \leq r \} \). If \( \theta_k \in B(r) \), then
\begin{align} \frac{\gamma_k}{n \exp(r/\lambda) + 1} I_n \preceq \Lambda_k \preceq \gamma_k I_n. 
\end{align}
\textbf{Smoothness of $L$.} The Hessian of $L$ has the following form 
\begin{align}
    \nabla^2 L := \begin{bmatrix}
        \nabla^2_{uu} L & \nabla^2_{uv} L \\ 
        \nabla^2_{vu} L & \nabla^2_{vv} L 
    \end{bmatrix} =  \begin{bmatrix}
 \diag(\sum_{i} M_{ij}) & M \\ 
  &  \\ 
 M^\top & \diag(\sum_{j}M_{ij})
    \end{bmatrix}
\end{align}
We will prove that the Hessian bounded within the domain $\theta \in B(r)$.
For all $v := \begin{bmatrix}
    s \in \R^{n} \\ 
    s' \in \R^n
\end{bmatrix}$ such that $\|v\|^2 = 1$, we have 
\begin{align}
    v^\top \nabla^2 L v = \|s\|^2_{\diag(\sum_{i}M_{ij})} + 2 s^\top M s' + \|s'\|^2_{  \diag(\sum_{j}M_{ij})} \label{eq:quadratic_Hessian}
\end{align}
where  \( \| v \|^2_A = v^\top A v \). Recall $M_{ij} = e^{\tfrac{-C_{ij}+u_i + v_j}{\lambda}-1}$, hence 
\begin{align}
    \sum_{ij}s_i s'_j M_{ij} & = \sum_{ij} s_i e^{\sfrac{u_i}{\lambda}} s_j e^{\sfrac{v_j}{\lambda}} e^{-\sfrac{C_{ij}}{\lambda}-1} \\ 
    & \leq \sum_{ij} s_i e^{\sfrac{u_i}{\lambda}} s_j e^{\sfrac{v_j}{\lambda}} \\
    & \leq \sqrt{\sum_i s_i^2 e^{2\sfrac{u_i}{\lambda}}} \sqrt{\sum_i (s'_i)^2 e^{2\sfrac{v_i}{\lambda}}}\\
    & \leq e^{\sfrac{2r}{\lambda}}
\end{align}
It is easy to check that $\|\diag(\sum_{i}M_{ij}) \|$ and $ \|\diag(\sum_{j}M_{ij}) \|$ are bounded by $n e^{\sfrac{r}{\lambda}}$. Replacing these inequalities into \eqref{eq:quadratic_Hessian} yields
\begin{align}
    v^\top \nabla^2 L v \leq n e^{\sfrac{r}{\lambda}} \left(\underbrace{\|s\|^2 + \|s'\|^2}_{=1}\right) + 2 e^{\sfrac{2r}{\lambda}} \leq (n+2) e^{\sfrac{2r}{\lambda}}.
\end{align}
Thus, $L(\theta)$ is $\zeta$-smooth for $\zeta: = (n+2) e^{\sfrac{2r}{\lambda}}$ when $\theta \in B(r)$.

\textbf{Boundedness of iterates.}  The recurrence relation of the iterates defined in \eqref{eq:gd} leads to the following inequality:
\begin{align} \label{eq:contraction}
    \| \theta_{k+1} - \theta^* \|^2_{\Lambda_k^{-1}} = \| \theta_{k} - \theta^* \|^2_{\Lambda_k^{-1}} - 2 \langle \theta_k - \theta^*, \nabla L(\theta_k) \rangle + \| \nabla L(\theta_k) \|^2_{\Lambda_k},
\end{align}
recall \( \| v \|^2_A = v^\top A v \).
Since \( L \) is \( \zeta \)-smooth within \( B(r) \), by Theorem 2.1.5 of \cite{nesterov2013introductory}, we have
\begin{align}
    \langle \nabla L(\theta), \theta - \theta^* \rangle \geq \frac{1}{\zeta} \| \nabla L(\theta) \|^2.
\end{align}  
Substituting the above inequality into \eqref{eq:contraction} yields
\begin{align} \label{eq:stepineq}
    \| \theta_{k+1} - \theta^* \|_{\Lambda_k^{-1}}^2 \leq \| \theta_{k} - \theta^* \|_{\Lambda_k^{-1}}^2 - \left( \frac{2}{\zeta} - \gamma_k \right) \| \nabla L(\theta_k) \|^2.
\end{align}  
Let \( \Delta_k := \| \theta_k - \theta^* \|^2_{\Lambda_k^{-1}} \). For \( \gamma_k = \frac{1}{\zeta} \), the above inequality ensures that \( \Delta_k \) is monotonically decreasing:
\[
\Delta_{k+1} \leq \Delta_k - \left( \frac{2}{\zeta} - \gamma_k \right) \| \nabla L(\theta_k) \|^2 \leq \Delta_k.
\]
To maintain \( \theta_k \in B(r) \) for all \( k \), choose \( r \) such that
\[
\| \theta_k \| \leq \Delta_k + \| \theta^* \|_{\Lambda_k^{-1}} \leq \| \Delta_1 \|_{\Lambda_1^{-1}} + \| \theta^* \|_{\Lambda_1^{-1}} \leq 2 \left( \| \theta_1 - \theta^* \| + \| \theta^* \| \right) = r.
\]
We now show that $\theta_k \in B(r)$ concludes the proof. 

\paragraph{Convergence to a stationary point.}  
Since $\theta_k \in B(r)$, we can take the  average of \eqref{eq:stepineq} over \( k = 1, \ldots, \ell \):
\begin{align*}
    \sum_{k=1}^\ell \| \nabla L(\theta_k) \|^2 \leq \zeta \left( \sum_{k=1}^\ell \Delta_k - \Delta_{\ell+1} \right) \leq  \zeta   \Delta_1 \leq \zeta\left( n e^{\sfrac{r}{\lambda}} + 1 \right) r.
\end{align*}  
The above inequality leads to the following bound on the minimum gradient norm:
\begin{align} \label{eq:grad_conv}
    \min_{k \leq \ell} \| \nabla L(\theta_k) \|^2 \leq \tfrac{1}{\ell} \sum_{k=1}^\ell \| \nabla L(\theta_k) \|^2  \leq \left(\frac{1}{\ell}\right)\zeta(n e^{\sfrac{r}{\lambda}}+1) r.
\end{align}

\textbf{Closeness to Doubly Stochastic Matrices.}  
By definition,
\begin{align}
    \nabla L(\theta_k) = \begin{bmatrix}
        M^{(k)} 1_n -  1_n \\
        (M^{(k)})^\top 1_n -  1_n
    \end{bmatrix},
\end{align}
where \( \mathbf{1} \) denotes the vector of all ones.  
Substituting the expression for \( \nabla L(\theta_k) \) into \eqref{eq:grad_conv} gives
\begin{align}
    \min_{k \leq \ell} \left( \| M^{(k)} 1_n  -1_n \|^2 + \| (M^{(k)})^\top 1_n -  1_n \|^2 \right) \leq \frac{\zeta(n \exp(r/\lambda) + 1) r}{\ell}.
\end{align}

\section{Proof of Proposition~\ref{prop:S}}
We prove $f(A) \in \S_{3\epsilon}$ and the proof for $g(A) \in \S_{3\epsilon}$ follows exactly the same.  
Since $A \in \S_\epsilon$, the following two inequalities hold 
\begin{align}
    \left|\sum_{i} A_{ij} - 1\right| \leq \epsilon \implies \sum_{i}A_{ij} \geq 1 - \left| 1 - \sum_{i} A_{ij}\right| \geq 1 - \epsilon
\end{align}
Using the above two inequalities, we proceed as
\begin{align}  |\tfrac{A_{ij}}{\sum_{i} A_{ij}} - A_{ij}| & = A_{ij} | 1 - \tfrac{1}{ \sum_{i}A_{ij}}| \\ 
& \leq \tfrac{A_{ij}\epsilon}{\sum_{i}A_{ij}} \\ 
& \leq \tfrac{A_{ij}\epsilon}{1-\epsilon}.
\end{align}
We use the above inequality to complete the proof:
\begin{align}
    \left|\sum_j \tfrac{A_{ij}}{\sum_i A_{ij}} - 1\right| & \leq \left|\sum_j \tfrac{A_{ij}}{\sum_i A_{ij}} - \sum_j A_{ij}\right|  + | \sum_{j} A_{ij} - 1|\\
    & \leq \sum_{j}|\tfrac{A_{ij}}{ \sum_{i} A_{ij}} - A_{ij}| + \epsilon \\ 
    & \leq \tfrac{\epsilon}{1-\epsilon}\sum_{j} A_{ij} + \epsilon \\ 
    & \leq \epsilon \left(1 + \tfrac{1+\epsilon}{1-\epsilon} \right) 
\end{align}
\section{Proof of Proposition~\ref{prop:d_bound}}

We prove part (i), and the proof for part (ii) follows exactly the same. The following inequality holds for $A \in \S_{\epsilon}$:
\begin{align}
    \forall j: \left| \sum_{i}A_{ij} - 1\right| \leq \epsilon. \label{eq:row_sum}
\end{align}
Using the above inequality, we get:
\begin{align}
    |q'_j - q_j| & = \left| \tfrac{q_j}{\sum_{i} A_{ij}} - q_j \right|  \\ 
    & = q_j | \frac{1}{\sum_{i} A_{ij}} -1 |  \\ 
    &\leq q_j \left( \tfrac{\epsilon}{1-\epsilon} \right)
\end{align}
Plugging the above inequality into $\mu$ concludes the proof: 
\begin{align}
    \frac{q_iq'_j}{q_j q'_i} \leq  \frac{1}{1-2\epsilon} \implies \mu(q,q'_i) \leq \log(\tfrac{1}{1-2\epsilon})\leq \tfrac{2\epsilon}{1-2\epsilon}.
\end{align}

\section{Supplementary experiments}

\begin{figure*}[t!]
    \centering
    \begin{tabular}{ M{3.5cm}   M{2.5cm} M{2.5cm} M{2.5cm} |M{3cm}}

& test $n=15$& &  &  optimal $P^*$ 
\\
(1) generalization & \includegraphics[width=0.12\textwidth]{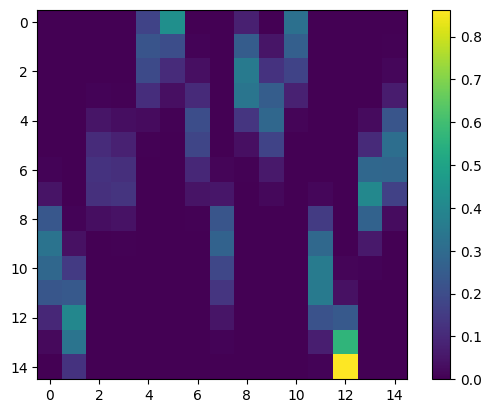} & &  & \includegraphics[width=0.12\textwidth]{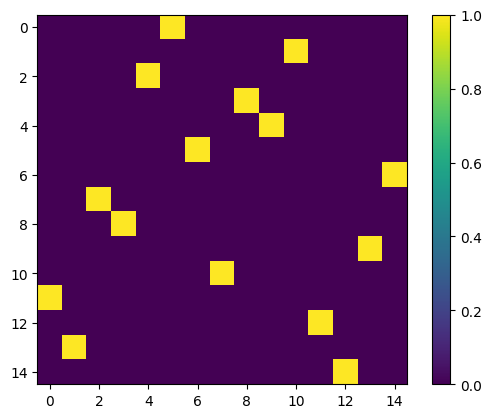}\\
&  
       without   & with 
& &   \\
(2) prompt engineering& \includegraphics[width=0.12\textwidth]{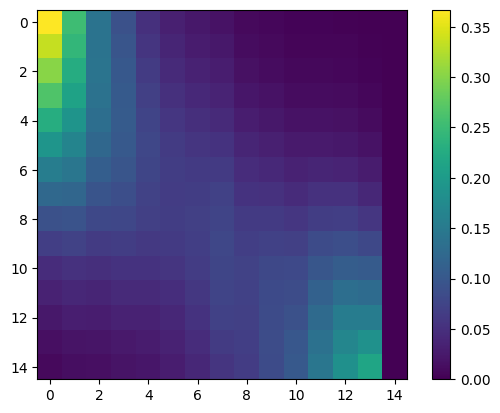}   & \includegraphics[width=0.12\textwidth]{images/last_15.png}   & &\includegraphics[width=0.12\textwidth]{images/ground_15.png} \\   

& layer 5 & layer 15 & last layer (20) &\\
(3) Iterative inference & \includegraphics[width=0.12\textwidth]{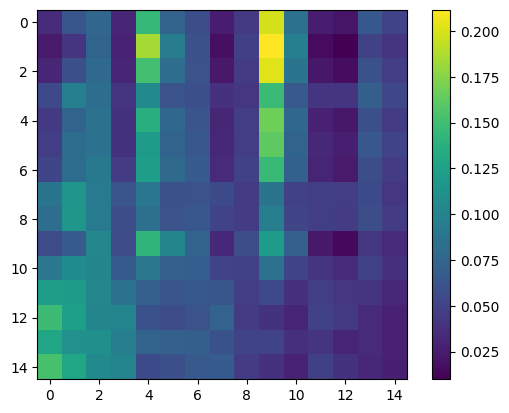}& \includegraphics[width=0.12\textwidth]{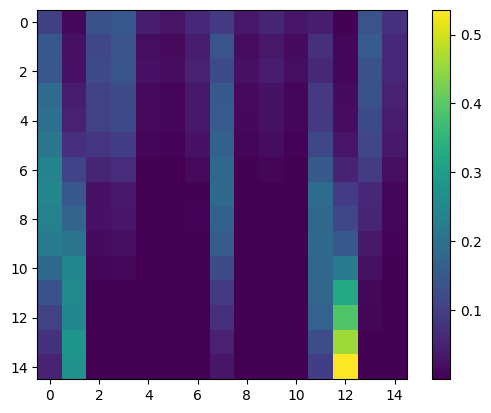} & \includegraphics[width=0.12\textwidth]{images/last_15.png}& \includegraphics[width=0.12\textwidth]{images/ground_15.png} 
\end{tabular}\caption{\footnotesize{\textbf{Observations on In-context Learning for OT.} We extend the experiments shown in Fig.~\ref{fig:train} to larger values of$n$.
\emph{(1)} The model is trained to solve OT with 7 data points and evaluated on 15 data points. 
The left image shows the attention weights, which closely approximate the OT solution shown on the right.
\emph{(2)} After specific prompt engineering, the attention weights between tokens estimate the OT solution. Notably, this prompt engineering is used in (1).
\emph{(3)} The attention weights evolve across layers, progressively yielding a more accurate approximation of the optimal solution. See Appendix~\ref{sec:experiments_app} for details.
}}
\label{fig:train_appendix}
\end{figure*}

Figure~\ref{fig:train_appendix} repeats the experiment in Figure~\ref{fig:train} for a larger $n$. 

\end{document}